\documentclass{report}
\usepackage{mathptmx}

\usepackage[a4paper]{geometry}
\usepackage{verbatim}
\usepackage{pdfpages}
\usepackage{setspace}
\usepackage{nomencl}
\usepackage{blindtext} 
\usepackage{amsmath} 
\usepackage{amssymb} 
\usepackage{array}

\usepackage{graphicx} 

\usepackage{longtable} 
\usepackage{bigstrut} 
\usepackage{enumerate} 
\usepackage{todonotes} 
\usepackage{makeidx} 
\makeindex

\usepackage{svg}
\usepackage{amsmath,amsthm,amssymb}
\usepackage{verbatim}
\usepackage{dsfont}
\usepackage[T1]{fontenc}
\usepackage[osf]{newpxtext}
\usepackage{textcomp} 
\usepackage[varg,bigdelims,cmintegrals,cmbraces]{newpxmath}
\usepackage[scr=rsfso]{mathalfa}
\usepackage{bm}
\usepackage{xspace}
\usepackage{soul} 

\usepackage{geometry}
\usepackage{setspace}

\usepackage{microtype}

\usepackage{natbib}
\usepackage{graphicx}
\usepackage{epstopdf}
\usepackage{epsfig}
\usepackage{color}
\usepackage{psfrag}
\usepackage{subfig}
\usepackage{tabulary}
\usepackage{tabularx}
\usepackage{booktabs}
\usepackage[flushleft]{threeparttable}
\usepackage{makecell}
\usepackage[title,titletoc,toc]{appendix}
\usepackage{placeins}
\usepackage{ragged2e}
\usepackage{caption}
\definecolor{darkblue}{rgb}{0,0,0.5}
\definecolor{mauve}{rgb}{0.88,0.69,1}
\usepackage[pdftex, colorlinks=true, citecolor=darkblue, linkcolor=darkblue, urlcolor=darkblue]{hyperref}
\hypersetup{
    pdftitle={},
    pdfauthor={},
    pdfsubject={},
    pdfkeywords={},
    bookmarksnumbered=true,     
    bookmarksopen=true,         
    bookmarksopenlevel=1,       
    colorlinks=true,            
    pdfstartview={XYZ null null 0.75},        
    pdfpagemode=UseOutlines,    
    pdfpagelayout=SinglePage
}
\usepackage{multirow}

\usepackage{titlesec}
\titleformat{\chapter}[display]
  {\normalfont\scshape\Large\centering}
  {\chaptertitlename\ \thechapter}{0pt}{\Huge}
\usepackage[bottom]{footmisc}
\usepackage[capposition=top]{floatrow}
\usepackage{lscape} 

\usepackage{rotating}

\usepackage{lineno,hyperref}
\modulolinenumbers[5]
\usepackage{amsmath}
\usepackage{amsfonts}
\usepackage{mathtools}
\usepackage{import}
\usepackage{graphicx}[width=\columnwidth] 
\usepackage[export]{adjustbox}
\usepackage{rotating}
\usepackage{amssymb}
\usepackage[inline]{enumitem}
\usepackage{multimedia}
\usepackage{graphics}
\usepackage{bbm}
\usepackage{pstricks}
\usepackage{pgf}
\usepackage{wrapfig} 
\usepackage{rotfloat}
\usepackage{float}
\usepackage{booktabs}
\usepackage{multirow}
\usepackage{pdflscape}

\providecommand{\printnomenclature}{\printglossary}
\providecommand{\makenomenclature}{\makeglossary}
\makenomenclature
\makeatletter

\providecommand{\LyX}{L\kern-.1667em\lower.25em\hbox{Y}\kern-.125emX\@}

\DeclarePairedDelimiterX{\kldivX}[2]{(}{)}{%
	#1\;\delimsize\|\;#2%
}

\usepackage{tauthesis}
\usepackage[font={small,bf}, labelfont={small,bf}, margin=1cm]{caption}
\usepackage{titlesec}
\newcommand{\hsp}{\hspace{20pt}}

\titleformat{\chapter}[hang]{\Huge\bfseries}{\thechapter\hsp}{0pt}{\Huge\bfseries}

\Title{\textbf{BiHRNN: Bi-Directional Hierarchical Recurrent Neural Network for Inflation Forecasting}}
\Author{Maya Vilenko}
\Year{December 2024}
\Supervisor{Prof. Noam Koeninstein}
\Department{School of Industrial Engineering}
\Degree{Master of Science}

\makeatother

\usepackage[english]{babel}
\begin{document}

\prelimpages

\titlepage

\chapter*{Abstract}

Inflation prediction is essential for guiding decisions on interest rates, investments, and wages, as well as for enabling central banks to establish effective monetary policies to ensure economic stability. The complexity of predicting inflation arises from the interplay of numerous dynamic factors and the hierarchical structure of the Consumer Price Index (CPI), which organizes goods and services into categories and subcategories to capture their contributions to overall inflation. This hierarchical nature demands advanced modeling techniques to achieve accurate forecasts.


In this work, we introduce Bi-directional Hierarchical Recurrent Neural Network (BiHRNN) model, a novel modeling approach that strikes a balance between these extremes by leveraging the hierarchical structure of datasets like the CPI. BiHRNN facilitates bidirectional information flow between hierarchical levels, where higher-level nodes influence lower-level ones and vice versa. This is achieved using hierarchical informative constraints applied to the parameters of Recurrent Neural Networks (RNNs), enhancing predictive accuracy across all hierarchy levels. By integrating hierarchical relationships without the inefficiencies of a unified model, BiHRNN offers an effective solution for accurate and scalable inflation forecasting.

We implemented our BiHRNN model on three distinct inflation datasets from major markets: the United States, Canada, and Norway, all of which include a variety of economic indices. For each use case, we gathered the necessary data, trained and evaluated the BiHRNN model, and fine-tuned its hyper-parameters. Additionally, we experimented with various loss functions to enhance the model's performance.

The results show that the BiHRNN model significantly outperforms traditional RNN approaches in forecasting accuracy across most levels of the hierarchy. The unique bidirectional architecture of the model, which facilitates the flow of information across hierarchical levels, played a crucial role in achieving these improvements.

Looking ahead, we aim to expand the application of BiHRNN to additional hierarchical inflation markets, exploring different domains, and refining the model to address the specific challenges of these datasets. This study provides a strong foundation for employing BiHRNN in inflation forecasting and underscores its potential to surpass traditional methods.

The code for this Thesis is available at:  \href{https://github.com/mayavilenko/Maya-Thesis}{https://github.com/mayavilenko/Maya-Thesis}.

\tableofcontents{}

\acknowledgments{I would like to express my deepest gratitude to my supervisor, Dr. Noam Koeningstein, for his invaluable guidance, support, and encouragement throughout this research. His expertise and insights have been instrumental in shaping the direction and outcomes of this work.}

\textpages

\listoffigures

\listoftables

\printnomenclature{}

\chapter{Introduction}
Inflation prediction is crucial for policymakers, businesses, and consumers as it influences decisions on interest rates, investment strategies, and wages. Accurate forecasts help central banks set appropriate monetary policies to maintain economic stability and control price levels. Businesses rely on inflation expectations to plan budgets, pricing strategies, and resource allocation, while consumers consider inflation trends when making purchasing and savings decisions.
However, inflation prediction is challenging due to the interplay of numerous dynamic factors that influence prices, such as monetary policy, supply chain disruptions, labor market conditions, and geopolitical events. Predicting inflation components, such as food, gas, and clothing, adds another layer of complexity, as these categories are affected by distinct factors like weather, global markets, and trade policies. Each component has unique drivers, making it difficult to aggregate these predictions into a comprehensive forecast.
Additionally, inflation expectations create feedback loops—when businesses and consumers anticipate rising prices, their actions, like increasing wages or raising prices, can directly contribute to higher inflation. These factors, combined with global and local economic interactions, make accurate inflation prediction a significant challenge.

The Consumer Price Index (CPI) is organized hierarchically, grouping goods and services into broad categories such as food, energy, and housing, which are further subdivided into detailed subcategories to capture their impact on overall inflation. This complex structure necessitates advanced modeling techniques.
There are several methods for training predictive models with hierarchical data. One approach is to train separate models for each series within the hierarchy, which can reduce the risk of under-fitting by focusing on specific data segments. However, this method often leads to overfitting due to the limited amount of data available for each individual model. An alternative  approach is to train a single model using all the hierarchical data combined, which can take advantage of larger datasets but tends to be computationally intensive and may struggle to capture the differing dynamics across various levels of the hierarchy.
Our approach, BiHRNN, achieves an effective middle ground by harnessing hierarchical relationships to enhance model performance, without the computational burden and inefficiency of training a single, unified model.

Bidirectional Hierarchical Recurrent Neural Networks (BiHRNN) are designed to model the hierarchical structure of datasets like the CPI, with each node in the network graph representing an RNN unit that captures the values of a specific (sub)-index. This architecture enables bidirectional information flow—allowing higher-level RNN nodes to influence lower-level ones and vice versa—through hierarchical informative priors applied to the RNNs' parameters. This bidirectional exchange enhances predictions across all levels of the hierarchy.

The BiHRNN represents a substantial advancement over its predecessor, the Hierarchical Recurrent Neural Network (HRNN), which allowed information to flow only in one direction—from parent categories to child categories. By enabling bidirectional communication, the BiHRNN enhances predictive accuracy and reliability. Evaluation results reveal that the BiHRNN consistently outperforms both the HRNN and traditional methods, providing more precise and dependable inflation predictions. Across various metrics, such as prediction accuracy, overall accuracy, correlations, and goodness of fit, the BiHRNN demonstrates notable improvements over the HRNN, underscoring its superior ability to capture and model the hierarchical dependencies within the data.

The central aim of this thesis is to present the BiHRNN model, specifically designed to utilize hierarchical data structures for enhanced prediction accuracy. 
We demonstrate the model's effectiveness using three distinct Consumer Price Index (CPI) datasets from Canada, Norway, and the US. In these cases, we forecast the CPI while accounting for multiple levels, ranging from broad economic indicators to specific categories, providing a comprehensive evaluation of the model's performance across different hierarchical structures.

The thesis is structured as follows:
Related work of existing methodologies in hierarchical data modeling and forecasting (pages 5-8), along with an overview of Recurrent Neural Networks (RNN) (page 9) and the previous Hierarchical Recurrent Neural Network (HRNN) model (pages 10-12).
A detailed explanation of the BiHRNN model, including its formulation and inference mechanisms (pages 12-16).
A description of the datasets used (pages 17-25), the baseline models (pages 26-27), evaluation metrics (page 28), and a thorough analysis of the BiHRNN model’s performance (pages 29-30).
Interpretation of the results and the implications of the findings (pages 30-32), and potential directions for future work (pages 33).

\chapter{Related Work}

In this section, we examine existing research on inflation data modeling and forecasting, with an emphasis on methodologies that make use of the hierarchical structure of data to address complex challenges effectively.

\section{Time series}
Time series prediction plays a crucial role in various domains, such as energy management, supply chain optimization, sports analytics, and weather forecasting. It involves forecasting future values by analyzing previously observed data in a sequential order. Traditional approaches such as Autoregressive Integrated Moving Average (ARIMA) models have been commonly applied in inflation forecasting due to their straightforward nature and ease of interpretation \citep{Box1970}. However, with the rise of deep learning techniques, more advanced models such as Recurrent Neural Networks (RNNs) and Long Short-Term Memory (LSTM) networks have demonstrated substantial improvements in predictive accuracy \citep{Hochreiter1997, Goodfellow2016}. Recent advancements in time series forecasting include the use of Transformers, as noted by \citep{zeng2022transformerseffectivetimeseries}. Transformers are highly effective at capturing complex patterns and long-range dependencies; however, their large number of parameters makes them susceptible to overfitting, particularly when applied to the smaller datasets often encountered in time series tasks. Additionally, their computational complexity, driven by the self-attention mechanism, can introduce unnecessary overhead, especially when simpler models often perform comparably for many forecasting tasks.

\subsection{Inflation Time-Series Prediction}
Forecasting inflation time series is critical for economic policy and decision-making. Traditionally, this has been achieved using econometric models such as Vector Autoregression (VAR) and structural models, which excel at capturing dynamic relationships between various economic indicators \citep{Sims1980}. However, recent advancements in time series forecasting have spurred increasing interest in deep learning techniques for inflation prediction. Models like LSTM networks and hybrid approaches combining statistical and machine learning methods have shown superior performance in capturing the complex and non-linear patterns inherent in inflation data \citep{Cheng2019, Guo2021}. 

The Filtered Ensemble Wavelet Neural Network (FEWNet) stands out for its innovative approach to forecasting CPI inflation \citep{SENGUPTA2024}. FEWNet utilizes wavelet transforms to decompose inflation data into high- and low-frequency components and integrates additional economic factors, such as economic policy uncertainty and geopolitical risk, to enhance forecasting accuracy. These transformed components, along with filtered exogenous variables, are processed through autoregressive neural networks to produce an ensemble forecast. By effectively capturing non-linearities and long-range dependencies through its adaptable architecture, FEWNet has demonstrated superior performance compared to traditional models, providing accurate forecasts and robust estimation of prediction uncertainty.

While some studies focus on machine learning techniques, other recent researches highlight the significance of trend and cross-sectional asymmetry measures in enhancing inflation forecasting methodologies. Trimmed-mean inflation estimators have demonstrated their effectiveness in predicting headline inflation for the Personal Consumption Expenditures (PCE) Price Index, substantially improving both point and density forecast accuracy over medium- and long-term horizons \citep{VERBRUGGE2024735}.

\section{Hierarchical inflation Forecasting}
Hierarchical inflation modeling structures inflation data into a hierarchy, where each disaggregated category, such as specific goods or services, contributes to the overall inflation index. This approach enhances prediction accuracy by leveraging the relationships and dependencies between different levels of the hierarchy, such as categories, subcategories, and aggregate indices. Forecasting within this framework often involves aggregating or disaggregating data to ensure consistency and coherence across all hierarchical levels, allowing for more precise and interpretable predictions.

An example of hierarchical inflation forecasting is the HRNN model \citep{BARKAN20231145}, the predecessor of BiHRNN. The HRNN model is specifically designed to address the unique challenges of inflation forecasting in datasets with hierarchical structures. Lower levels, such as specific categories of goods or regional inflation indices, often exhibit missing data and higher volatility in price changes compared to higher aggregate levels. By aligning with the principles of hierarchical Bayesian models, the HRNN assigns prior distributions to parameters at each level, capturing the relationships and dependencies within the hierarchy. 

Hierarchical time series inflation models are particularly well-suited for inflation forecasting as they effectively integrate temporal dynamics and cross-level information. This capability is crucial for tasks like predicting the consumer price index, where capturing temporal trends, seasonality, and structural shifts in economic conditions is essential. For instance, fluctuations in specific product categories or regional price indices can influence aggregate inflation measures, making it important to account for these interactions. By leveraging information across hierarchical levels, these models produce robust forecasts that address both short-term volatility and long-term trends. Building on this foundation, we chose to explore another hierarchical model that enables bidirectional information propagation, further enhancing prediction accuracy and capturing more intricate relationships within the data.

Conventional hierarchical forecasting methods generally use top-down, bottom-up, or middle-out strategies. In the top-down approach, forecasts are generated at the highest level of the hierarchy and then allocated to lower levels. In contrast, the bottom-up approach aggregates forecasts from the lowest level to produce predictions for higher levels. The middle-out approach combines elements of both top-down and bottom-up methods \citep{Hyndman2011}.

Recent advancements have incorporated machine learning techniques into hierarchical forecasting. Models like hierarchical RNNs (HRNNs) and hierarchical LSTMs exploit the hierarchical structure of data to capture dependencies both within and across different levels. These models have demonstrated improved performance across various applications, such as sales forecasting and energy demand prediction \citep{Wickramasuriya2019}. By leveraging information from multiple levels of the hierarchy, they offer more accurate and reliable predictions compared to traditional non-hierarchical approaches.

\section{Hierarchical Long Short-Term Memory Network}
Hierarchical Long Short-Term Memory networks (Hierarchical LSTMs) are an extension of traditional LSTM models, designed to handle data with multiple levels of structure, such as hierarchical time series or sequential data organized into nested groups. These models capture dependencies both within and across different levels of the hierarchy by processing information at each level of granularity. For example, in time series forecasting, a hierarchical LSTM can model trends at higher levels (e.g., overall sales) while simultaneously capturing detailed patterns at lower levels (e.g., sales by region or product category). This architecture allows the model to account for complex interactions across levels, improving prediction accuracy and robustness in tasks where hierarchical relationships are important \citep{LIN2022107618}.

\section{Hierarchical Attention Network}
Hierarchical Attention Networks (HANs) are a deep learning framework designed to model hierarchical structures within data, particularly useful in natural language processing tasks. Unlike standard attention mechanisms, HANs apply attention at different levels of the data hierarchy, such as words within sentences and sentences within documents. This enables the model to focus on the most relevant components at each level, improving its ability to capture contextual relationships and interactions across the hierarchy. For example, in document classification, HANs can first highlight key words within individual sentences, then identify the most important sentences across the document. This hierarchical attention approach not only boosts interpretability but also enhances performance in tasks that rely on understanding multi-level data structures. The effectiveness of HANs has made them essential in various NLP applications, including document classification, sentiment analysis, and text summarization \citep{han_paper}.

\chapter{BiHRNN: Bi-Directional Hierarchical Recurrent Neural Network}

Hierarchical data refers to structured data organized into multiple levels, with each level representing a different degree of aggregation or detail. In the context of inflation, hierarchical data can range from the overall inflation index, such as the Consumer Price Index (CPI), down to disaggregated components like regional indices or individual product categories. This multi-level structure allows for analysis at varying levels of granularity, capturing relationships and dependencies across different layers. Such an approach is particularly valuable for inflation modeling, as changes at lower levels, such as specific goods or services, can propagate upward and influence aggregate measures, enabling more accurate and comprehensive forecasting.

\section{Hierarchical Recurrent Neural Networks}
Before delving into the specifics of the BiHRNN model, we provide a brief overview of its predecessor, the HRNN model \citep{BARKAN20231145}. The HRNN model is specifically designed to address the challenges of inflation forecasting in hierarchically structured datasets, where lower levels are often characterized by data sparsity and heightened volatility in change rates. To enhance predictions, the HRNN propagates information from parent categories to node categories within the hierarchy by employing hierarchical Gaussian priors. This approach connects each node’s parameters to its parent’s, allowing the model to share information across levels of aggregation. By leveraging parent-level information, the HRNN mitigates the effects of sparse or noisy data at finer levels and ensures consistency in forecasts across the hierarchy. Furthermore, it utilizes RNNs, specifically Gated Recurrent Units (GRUs) \citep{GRU}, which incorporate a feedback loop, enabling predictions to account for temporal dependencies. This combination of hierarchical priors and temporal modeling makes the HRNN particularly effective for capturing both cross-level interactions and dynamic patterns in inflation data \citep{RNNs_Book, RNN_evaluations}.

\subsection{Gated Recurrent Units (GRU)}
A Gated Recurrent Unit (GRU) is a type of Recurrent Neural Network (RNN) designed to capture long-term dependencies in sequential data by addressing the vanishing gradient problem common in traditional RNNs. GRUs use two gates—update and reset—to manage information flow. The update gate controls how much past information is passed along to future states, while the reset gate determines how much past information is forgotten, allowing GRUs to retain relevant information over longer sequences.

\begin{figure}[!htb]
    \centering
    {   \makebox[\textwidth]{
        \includegraphics[scale=0.6]{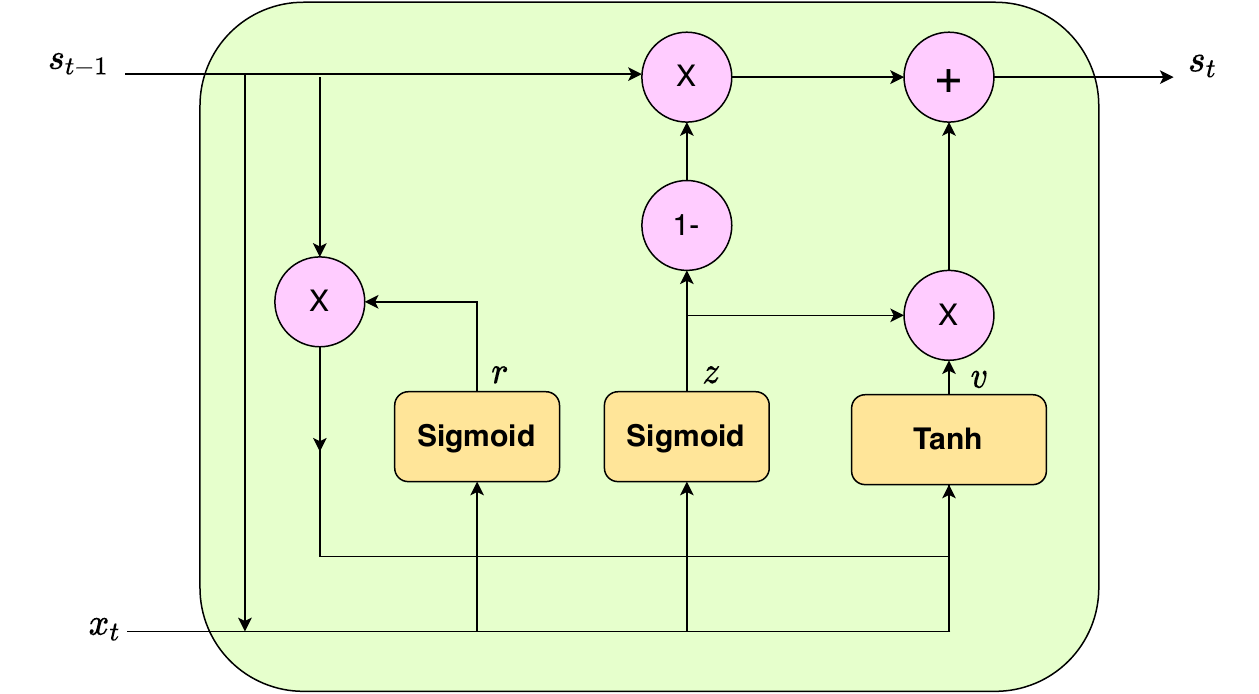}}
    }
    \caption
    {An illustration of a GRU unit.}
    \label{fig:GRU}
     \floatfoot*{\textit{Each line represents a vector, connecting the output of one node to the inputs of others. Pink circles indicate point-wise operations, while yellow boxes represent learned neural network layers. When lines merge, it signifies concatenation; when a line forks, it means the vector is copied, with each copy directed to different destinations.}}
\end{figure}

The following set of equations defines a GRU unit: 
\begin{equation}
\label{eq:gru}
\begin{split}
    z = & \sigma(x_tu^z + s_{t-1}w^z + b^z), \\
    r = & \sigma(x_tu^r + s_{t-1}w^r + b^r), \\
    v = & \tanh{(x_tu^v+(s_{t-1} \times r)w^v +b^v)}, \\
    s_t = & z \times v + (1-z)s_{t-1},
\end{split}
\end{equation}
where $u^z$, $w^z$ and $b^z$ are learned parameters governing the \emph{update gate} $z$, while $u^r$, $w^r$ and $b^r$ are the learned parameters for the \emph{reset gate} $r$. The candidate activation $v$ determined by the input $x_t$ and the previous output $s_{t-1}$, and is influenced by the learned parameters: $u^v$, $w^v$ and $b^v$. Finally, the output $s_t$ is a combination of the candidate activation $v$ and the previous state $s_{t-1}$ controlled by the \emph{update gate} $z$. Figure~\ref{fig:GRU} depicts an illustration of a GRU unit.

GRUs form the foundational unit of the HRNN model, detailed in Section \ref{subsec:hrnn} as well as our BiHRNN model detailed in Section \ref{subsec:model}.

\subsection{HRNN}\label{subsec:hrnn}
Next, we proceed with a description of the HRNN model. 
The following notations apply to both HRNN and Bidirectional HRNN models:

Let \(\mathcal{I}=\{n\}_{n=1}^N\) represent dataset graph nodes, each associated with a parent \(\pi_n\). For node \(n\), \(x_t^n\) is its observed value at time \(t\), and \(X_t^n\) represents the sequence up to \(t\).
A parametric function \(g\) (a GRU node) predicts the next value in the sequence, learning parameters \(\theta_n\) to predict \(x_{t+1}^n\). Assuming Gaussian errors, the likelihood of the time series is modeled as a product of normal distributions, with \(\tau_n^{-1}\) as the error variance.
A hierarchical informative prior connects each node’s parameters to its parent’s, with the prior \( p(\theta_n|\theta_{\pi_n},\tau_{\theta_n}) \) using \(\tau_{\theta_n}\) as a precision parameter. Higher \(\tau_{\theta_n}\) values indicate stronger parameter connections between node \(n\) and its parent \(\pi_n\).
Instead of globally optimizing \(\tau_{\theta_n}\), the HRNN sets \(\tau_{\theta_n} = e^{\alpha + C_n}\), where \(\alpha\) is a hyperparameter and \(C_n\) is the Pearson correlation between node \(n\) and its parent \(\pi_n\). This ensures that node \(n\) stays close to \(\pi_n\) in parameter space, especially when correlation is high. For the root node, a non-informative Gaussian prior with zero mean and unit variance is used.

Let \(X=\{X_{T_n}^n\}_{n\in \mathcal{I}}\) represent all time series, \(\theta=\{\theta_n\}_{n\in \mathcal{I}}\) the GRU parameters, and \(\tau=\{\tau_n\}_{n\in \mathcal{I}}\) the precision parameters. Here, \(X\) is observed, \(\theta\) contains learned variables, and \(\tau\) is defined by \(\tau_{\theta_n}\).

Given the the likelihood of the observed time series and the priors aforementioned above, the posterior probability is then extracted and is formulated according to Equation~\eqref{eq:posterior}.
\begin{equation}
\label{eq:posterior}
\begin{split}
p(\theta|X,\tau) &= \frac{p(X|\theta,\tau)p(\theta)}{P(X)} \propto  
\prod_{n\in \mathcal{I}}\prod_{t=1}^{T_n}\mathcal{N}(x_t^n;g(\theta_n,X_{t-1}^n),\tau_n^{-1})\prod_{n\in \mathcal{I}}\mathcal{N}(\theta_n;\theta_{\pi_n},\tau_{\theta_n}^{-1}\mathbf{I}).
\end{split}
\end{equation}

HRNN optimization follows a \textit{Maximum A-Posteriori} (MAP) approach to find the optimal parameters \(\theta^*\) by maximizing the posterior probability.
\begin{equation}
\label{eq:obj}
\theta^*=\underset{ \theta}{\text{argmax}}\log p(\theta|X,\tau).
\end{equation}

\begin{figure}[H]
    \centering
    \includegraphics[scale=0.25]{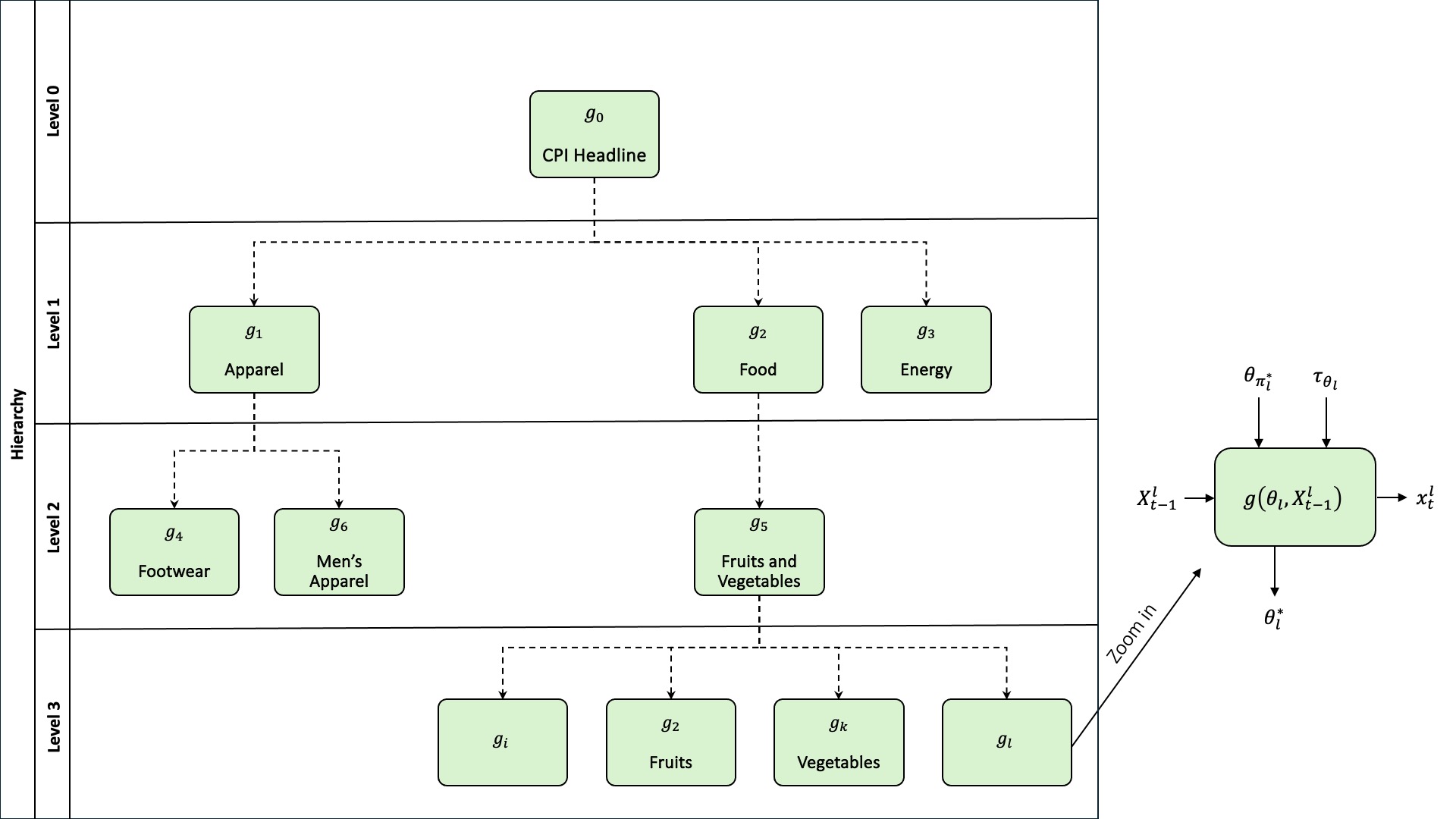}
    \caption{Illustration of the HRNN Model.}
    \label{fig:HRNN}
    
\end{figure}

The optimization is performed using stochastic gradient ascent on this objective. Figure~\ref{fig:HRNN} illustrates the HRNN architecture.

\section{Bidirectional HRNN Model}\label{subsec:model}

The Bidirectional HRNN (BiHRNN) builds upon the HRNN framework by enabling bidirectional information flow within hierarchical data structures, addressing a critical limitation of its predecessor. While the HRNN model effectively propagates information from parent categories to child categories, improving predictions for lower, more volatile levels, it does not leverage the potential benefits of propagating information in the reverse direction—from child categories back to their parents. Granular-level data often contains unique patterns or anomalies that can inform and refine higher-level predictions. By incorporating  bidirectional information flow, BiHRNN enhances the consistency and accuracy of predictions across all levels of the hierarchy.

The motivation for this extension lies in the interconnected nature of hierarchical data, particularly in the context of inflation forecasting. Economic indices at higher aggregation levels, such as national inflation rates, are directly influenced by fluctuations in disaggregated categories, such as specific goods, services, or regional indices. Without accounting for these influences, predictions at higher levels may miss critical insights embedded in lower-level data. By enabling information to flow upward, BiHRNN allows parent-level categories to benefit from the granularity and detail captured at the child level, thereby improving overall forecast accuracy and coherence across the entire hierarchy.

BiHRNN introduces a dual-constraint formulation, which integrates both top-down and bottom-up information; the constraints are applied during training to ensure structural consistency; and the enhanced loss function that governs its optimization, balancing accuracy across all hierarchical levels. These advancements make BiHRNN a more robust and versatile tool for forecasting in complex hierarchical datasets, particularly in domains like inflation modeling where cross-level interactions are critical.

\subsection{Bidirectional Information Flow}

BiHRNN is formulated as a risk minimization optimization problem. Instead of HRNN's informative prior, BiHRNN introduces two constraints on the models parameters. One ties the parameters of each time series to its parent's (similar to HRNN's prior) and the second ties the parameters of each of its' child series, with an appropriate weight:
\begin{itemize}
    \item \textbf{Parent-Node Constraint:} This constraint governs the relationship between the parameters of a node $n$ and its parent $\pi_n$. By aligning the parameters of node $n$ with those of its parent, this constraint ensures hierarchical consistency and allows top-down information to flow through the network.
    \item \textbf{Child-Node Constraint:} This constraint governs the relationship between node $n$ and its children $\eta_{i_n}$, where $\eta_{i_n}$ represents the $i$-th child of node $n$. This enables the node to aggregate information from its children, allowing bottom-up feedback to influence higher-level nodes.
\end{itemize}

By incorporating these two constraints, BiHRNN enables information to flow in both directions---\textit{downward} from parent to child and \textit{upward} from child to parent. This bidirectional approach significantly enhances the model’s ability to capture complex dependencies and interactions across hierarchical levels, leading to improved predictive performance.

\subsection{Customized Loss Function}

The information flow in BiHRNN is governed by a \textit{customized loss function} that balances prediction accuracy with hierarchical consistency. The loss function comprises three key components:

\begin{enumerate}
    \item \textbf{Mean Squared Error (MSE):}
    The primary objective of BiHRNN is to minimize prediction errors. This is achieved using the mean squared error:
    \begin{equation}
    \text{MSE} = \frac{1}{N} \sum \left( y - \hat{y} \right)^2
    \end{equation}
    where $y$ represents the observed values, $\hat{y}$ denotes the predicted values, and $N$ is the number of observations.
    
    \item \textbf{Parent Regularization ($l_{\text{parent}}$):}
    To ensure hierarchical coherence, the model penalizes the squared Euclidean distance between the parameters of a node $n$ and its parent $\pi_n$:
    \begin{equation}
    l_{\text{parent}} = \left( \theta_{\text{parent}} - \theta \right)^2
    \end{equation}
    This term enforces consistency between a node and its parent, ensuring that higher-level nodes influence lower-level nodes appropriately.
    
    \item \textbf{Child Regularization ($l_{\text{child}}$):}
    Similarly, the model incorporates the influence of child nodes through a weighted penalty:
    \begin{equation}
    l_{\text{child}} = \sum_{i \in \text{children}} w_i \left( \theta - \theta_{\text{child}} \right)^2
    \end{equation}
    where $w_i$ is a weight controlling the contribution of each child. This term allows the parameters of node $n$ to aggregate information from its children, enabling bottom-up information flow.
\end{enumerate}

The final loss function combines these components, weighted by hyperparameters $\lambda_1$ and $\lambda_2$ to control the relative importance of parent and child regularization:
\begin{equation}
\text{Loss}_{\text{BiHRNN}} = \frac{1}{N} \sum \left( y - \hat{y} \right)^2 
+ \lambda_1 \cdot l_{\text{parent}} + \lambda_2 \cdot l_{\text{child}}
\end{equation}

\subsection{Hyperparameter Tuning}

The hyperparameters $\lambda_1$ and $\lambda_2$ play a critical role in balancing the bidirectional information flow:
\begin{itemize}
    \item A higher $\lambda_1$ emphasizes top-down influence by prioritizing alignment with parent nodes.
    \item A higher $\lambda_2$ strengthens bottom-up feedback from child nodes.
\end{itemize}

Proper tuning of these hyperparameters is essential for optimizing the model’s performance while maintaining hierarchical consistency. To achieve this, Optuna\footnote{\url{https://optuna.org/}} was employed for hyperparameter tuning, utilizing its Tree-structured Parzen Estimator (TPE), a Bayesian optimization approach, to efficiently explore the hyperparameter space and ensure robust performance.

\subsection{Fixed Constraints During Training}

A key feature of the BiHRNN model is its use of \textit{fixed constraints} for the parent and child relationships throughout the training process. The procedure involves the following steps:
\begin{enumerate}
    \item \textbf{Pretraining the Base Model:} The HRNN (or a similar baseline model) is first trained independently for each category, and the learned weights are saved. These weights serve as the initial representations of the parent and child nodes.
    \item \textbf{Freezing the Weights:} Once the HRNN weights are trained, they are frozen and used as fixed constraints for the bidirectional model. This means that the weights representing parent and child relationships remain constant during the BiHRNN training process.
    \item \textbf{Stabilization and Regularization:} By leveraging frozen weights, the BiHRNN anchors predictions, ensuring that hierarchical relationships are preserved and reducing the risk of overfitting. This approach is particularly beneficial for datasets with limited samples or high variability, as it provides a stable foundation for the bidirectional model.
\end{enumerate}

\subsection{Summary}

The BiHRNN introduces a bidirectional approach to hierarchical modeling, leveraging dual constraints and a customized loss function to enable efficient information flow between nodes. By using fixed constraints and balancing top-down and bottom-up interactions, the model achieves superior forecasting accuracy and hierarchical coherence, establishing itself as a robust framework for hierarchical time series forecasting.

Figure ~\ref{fig:Bidirectional HRNN} below depicts the BiHRNN architechture.

\begin{figure}[H]
    \centering
    {   \makebox[\textwidth]{
        \includegraphics[scale=0.25]{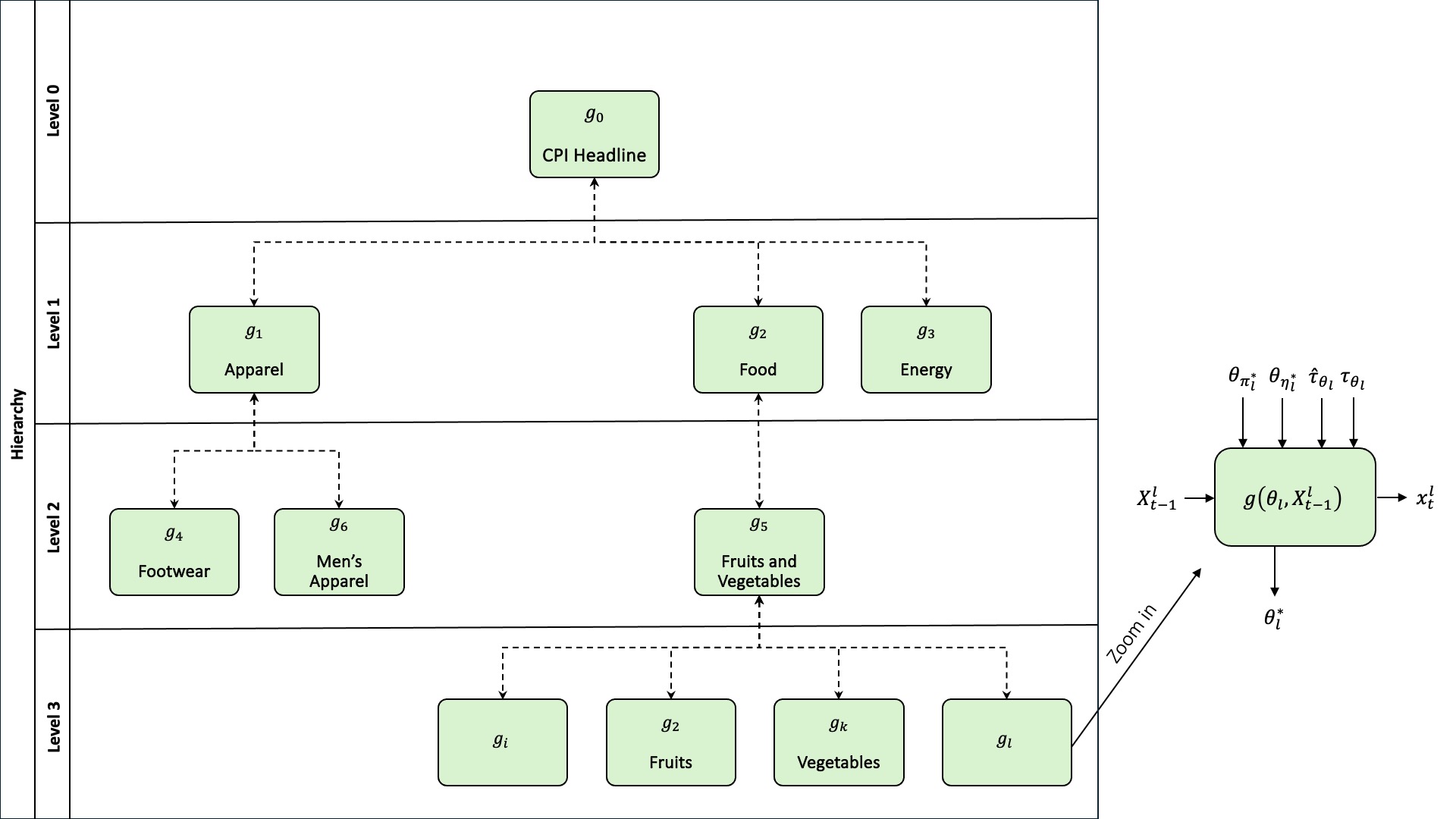}}
    }
        \caption
    {An illustration of our  BiHRNN Model.}
    \label{fig:Bidirectional HRNN}
\end{figure}

\chapter{Dataset} 
This work is based on monthly CPI data released by the US Bureau of Labor and Statistics (BLS) \footnote{Taken from \url{www.bls.gov/cpi/overview.htm}.}, Statistics Canada \footnote{Taken from \url{www.statcan.gc.ca/en/start}.}, and Statistics Norway \footnote{Taken from \url{www.ssb.no/en}.}, covering different time periods: the US dataset spans from January 2012 to September 2023, the Norway dataset from January 2009 to May 2023, and the Canada dataset from January 2013 to February 2023. In the following section, we outline the dataset's characteristics and detail our pre-processing methodology. To ensure reproducibility, the final version of the processed data is included within our BiHRNN code.

\section{The US Consumer Price Index}

The CPI for each month is released by the BLS a few days into the following month. Price data is gathered from around 24,000 retail and service establishments across 75 urban areas in the US. Additionally, housing and rent information is collected from approximately 50,000 landlords and tenants nationwide.
The BLS provides two distinct CPI measurements based on urban demographics:
\begin{enumerate} \item
    The {\bf CPI-U}, or Consumer Price Index for All Urban Consumers, covers about 93\%  of the total population. The items included in the CPI, along with their relative weights, are determined based on the Consumer Expenditure Survey, which estimates household spending. These items and weights are updated annually each January.
    \item 
    The {\bf CPI-W}, or Consumer Price Index for Urban Wage Earners and Clerical Workers, covers around 29\% of the population. This index focuses on households where at least 50\% of income is earned through wage-paying or clerical jobs, with at least one household member employed for 70\% or more of the year. The CPI-W is commonly used to track changes in benefit costs and inform future contract obligations.
\end{enumerate}    

In this work, we focus on the CPI-U, as it is widely regarded as the most reliable indicator of the average cost of living in the United States. The CPI-U prices are seasonally adjusted, with updates released in February that reflect price movements from the previous calendar year. It is important to note that, over time, new indexes have been introduced while others have been discontinued, causing shifts in the hierarchical structure of the dataset, which adds complexity to our analysis.

\subsection{US CPI Hierarchy}
The CPI-U is structured as an nine-level hierarchy consisting of 350 distinct nodes (indexes). At the top, Level 0 represents the headline CPI, which is the aggregated index of all components. Each index in the hierarchy is assigned a weight between 0 and 100, reflecting its contribution to the headline CPI at Level 0. 

Level 1 contains the 8 (not including ``All items excluding X'' categories) main aggregate categories, or sectors: (1) ``Food and Beverages,'' (2) ``Housing,'' (3) ``Apparel,'' (4) ``Transportation,'' (5) ``Medical Care,'' (6) ``Recreation,'' (7) ``Education and Communication,'' and (8) ``Other Goods and Services.''

Mid-levels (2-5) include more specific groupings such as ``Milk'' and ``Motor Fuel.'' The lower levels (6-8) feature finer-grained indexes, such as ``Rice,'' ``Child Care and Nursery School,'' ``Legal Services,'' ``Funeral Expenses,'' and ``Parking Fees and Tolls,'' among others. 

\section{Canada Consumer Price Index}

The  CPI in Canada is measured on a monthly basis by Statistics Canada\footnote{\href{https://www.statcan.gc.ca/en/start}{https://www.statcan.gc.ca/en/start}}. 
The target population for the CPI includes families and individuals living in urban and rural private households, excluding those in communal or institutional settings like prisons or long-term care. The price sample is gathered from various geographic areas, goods, services, and retail outlets to estimate price changes. Outlet selection focuses on high-revenue retailers, with prices mainly collected from retail stores and agencies. The relative importance of items in the CPI basket is derived using data from the Household Final Consumption Expenditure (HFCE) and the Survey of Household Spending (SHS).
It should be noted that in some years, category weights for the CPI were unavailable. Therefore, we employed Linear Regression to estimate the missing weights, with the CPI values of the child categories serving as independent variables and the parent category CPI as the dependent variable. We will expand on this further in Section \ref{subsec:data_aug}.

\subsection{Canada CPI Hierarchy}

The CPI is organized into a seven-level hierarchy with 293 distinct nodes (indexes). At the top, Level 0 represents the headline CPI, an aggregated index of all components. Each index in the hierarchy is assigned a weight between 0 and 1, indicating its contribution to its parent CPI.

Level 1 consists of 10 (not including ``All items excluding X'' categories) main aggregate categories, or sectors: (1) ``Food,'' (2) ``Shelter,'' (3) ``Household operations, furnishings and equipment,'' (4) ``Clothing and footwear,'' (5) ``Transportation,'' (6) ``Health and personal care,'' (7) ``Recreation, education and reading,'' (8) ``Alcoholic beverages, tobacco products and recreational cannabis,'' (9) ``Food and energy,'' and (10) ``Energy.''

Mid-levels (2-4) include more specific categories like ``Meat'' and ``Women's clothing.'' The lower levels (5-6) feature more detailed indexes such as ``Whole milk,'' ``Pasta mixes,'' ``Purchase of used passenger vehicles,'' and ``Roasted or ground coffee,'' alongside others.

\section{Norway Consumer Price Index}

The CPI in Norway is collected monthly by Statistics Norway\footnote{\href{https://www.ssb.no/en}{https://www.ssb.no/en}}, primarily through electronic questionnaires sent to outlets on the 10th of each month, with responses due by the first working day after the 15th. In addition, electronic scanner data from grocery stores, pharmacies, clothing, sports retailers, and petrol stations are received monthly. The sample includes around 650 goods and services, selected based on household budget surveys and industry information, covering approximately 2,000 firms. For rent surveys, 2,500 tenants are sampled from the Rental Market Survey. Firms are chosen from the Business Register, with larger firms having a higher probability of selection based on turnover, after stratifying by industry and region.

\subsection{Norway CPI Hierarchy}
The CPI is arranged in a three-level hierarchy with 52 distinct nodes (indexes). At the top, Level 0 represents the headline CPI, which aggregates all components. Each index in the hierarchy is given a weight between 0 and 1000, reflecting its contribution to the overall CPI.

Level 1 is made up of 12 main aggregate categories, or sectors: (1) ``Food and non-alcoholic beverages,'' (2) ``Alcoholic beverages and tobacco,'' (3) ``Clothing and footwear,'' (4) ``Housing, water, electricity, gas and other fuels,'' (5) ``Furnishings, household equipment and routine maintenance,'' (6) ``Health,'' (7) ``Transport,'' (8) ``Communications,'' (9) ``Recreation and culture,'' (10) ``Education,'' (11) ``Restaurants and hotels,'' and (12) ``Miscellaneous goods and services.''

Level 2 includes more specific categories like ``Footware'' and ``Clothing.''

The tables below hold data of the first three hierarchies of the US CPI (levels 0-2):

\setlength{\tabcolsep}{2pt}
\renewcommand{\arraystretch}{0.9}

\begin{table}[H]
\caption{Indexes Level 0 And 1 - US} \label{tab:indexes_level_0_1}

\begin{center}
{\small
\begin{tabularx}{\textwidth}{@{}Xll@{}}
\toprule
Level & Index & Parent \\ 
\midrule
0 & All items & - \\ 
\midrule
1 & All items less energy & All items \\ 
1 & All items less food & All items \\ 
1 & All items less food and energy & All items \\ 
1 & All items less food and shelter & All items \\ 
1 & All items less food, shelter, and energy & All items \\ 
1 & All items less food, shelter, energy, and used cars and trucks & All items \\ 
1 & All items less homeowners costs & All items \\ 
1 & All items less medical care & All items \\ 
1 & All items less shelter & All items \\ 
1 & Apparel & All items \\ 
1 & Apparel less footwear & All items \\ 
1 & Commodities & All items \\ 
1 & Commodities less food & All items \\ 
1 & Durables & All items \\ 
1 & Education and communication & All items \\ 
1 & Energy & All items \\ 
1 & Entertainment & All items \\ 
1 & Food & All items \\ 
1 & Food and beverages & All items \\ 
1 & Fuels and utilities & All items \\ 
1 & Household furnishings and operations & All items \\ 
1 & Housing & All items \\ 
1 & Medical care & All items \\ 
1 & Nondurables & All items \\ 
1 & Nondurables less food & All items \\ 
1 & Nondurables less food and apparel & All items \\ 
1 & Other goods and services & All items \\ 
1 & Other services & All items \\ 
1 & Recreation & All items \\ 
1 & Services & All items \\ 
1 & Services less medical care services & All items \\ 
1 & Services less rent of shelter & All items \\ 
1 & Transportation & All items \\ 
1 & Utilities and public transportation & All items \\ \bottomrule
\end{tabularx}}
\floatfoot*{\textit{Note}: Levels and Parents of Indexes might change through time}
\end{center}
\end{table}

\begin{table}[H]
\renewcommand{\arraystretch}{1}

\caption{Indexes Level 2 - US} \label{tab:indexes_level_2}
\begin{center}
{\footnotesize
\begin{tabularx}{\textwidth}{@{}Xll@{}}
\toprule
Level & Index & Parent \\ \midrule
2 & All items less food and energy & All items less energy \\ 
2 & Apparel commodities & Apparel \\ 
2 & Apparel services & Apparel \\ 
2 & Commodities less food & Commodities \\ 
2 & Commodities less food and beverages & Commodities \\ 
2 & Commodities less food and energy commodities & All items less food and energy \\ 
2 & Commodities less food, energy, and used cars and trucks & Commodities \\ 
2 & Communication & Education and communication \\ 
2 & Domestically produced farm food & Food and beverages \\ 
2 & Education & Education and communication \\ 
2 & Energy commodities & Energy \\ 
2 & Energy services & Energy \\ 
2 & Entertainment commodities & Entertainment \\ 
2 & Entertainment services & Entertainment \\ 
2 & Food & Food and beverages \\ 
2 & Food at home & Food \\ 
2 & Food away from home & Food \\ 
2 & Footwear & Apparel \\ 
2 & Fuels and utilities & Housing \\ 
2 & Homeowners costs & Housing \\ 
2 & Household energy & Fuels and utilities \\ 
2 & Household furnishings and operations & Housing \\ 
2 & Infants’ and toddlers’ apparel & Apparel \\ 
2 & Medical care commodities & Medical care \\ 
2 & Medical care services & Medical care \\ 
2 & Men’s and boys’ apparel & Apparel \\ 
2 & Nondurables less food & Nondurables \\ 
2 & Nondurables less food and apparel & Nondurables \\ 
2 & Nondurables less food and beverages & Nondurables \\ 
2 & Nondurables less food, beverages, and apparel & Nondurables \\ 
2 & Other services & Services \\ 
2 & Personal and educational expenses & Other goods and services \\ 
2 & Personal care & Other goods and services \\ 
2 & Pets, pet products and services & Recreation \\ 
2 & Photography & Recreation \\ 
2 & Private transportation & Transportation \\ 
2 & Public transportation & Transportation \\ 
2 & Rent of shelter & Services \\ 
2 & Services less energy services & All items less food and energy \\ 
2 & Services less medical care services & Services \\ 
2 & Services less rent of shelter & Services \\ 
2 & Shelter & Housing \\ 
2 & Tobacco and smoking products & Other goods and services \\ 
2 & Transportation services & Services \\ 
2 & Video and audio & Recreation \\ 
2 & Women’s and girls’ apparel & Apparel \\ \bottomrule
\end{tabularx}}
\floatfoot*{\textit{Note}: Levels and Parents of Indexes have changed over the years.}
\end{center}
\end{table}
\clearpage


\section{Data Preparation}\label{subsec:data_prep}
The hierarchical CPI data is provided as monthly index values. We transformed these CPI values into monthly logarithmic change rates. Let \(x_{t}\) represent the CPI value (of any node) at month \(t\). The logarithmic change rate at month \(t\), denoted as \(rate(t)\), is calculated as follows:
\begin{equation}
\label{eq:ratio}
rate(t) = 100 \times \log \left(\frac{x_{t}}{x_{t-1}}\right).
\end{equation}

Unless specified otherwise, the remainder of this paper focuses on monthly logarithmic change rates as defined in Equation~\eqref{eq:ratio}.

We divided the data into a \emph{training} set and a \emph{test} set as follows: for each time series, the first 75\% of the measurements (earliest in time) were assigned to the \emph{training} set, while the remaining 25\% were set aside for the \emph{test} set. The \emph{training} set was used to train the BiHRNN model and other baseline models. The \emph{test} set was used for evaluation. The results presented in Section~\ref{sec:Results} are based on this data split.

\subsection{General Data Statistics}
Table~\ref{tab:general-statistics} summarizes the number of data points and general statistics of the CPI time series after applying Equation~\eqref{eq:ratio}. A comparison between the headline CPI and the full hierarchy reveals that lower levels exhibit significantly higher standard deviations (STD) and wider ranges, indicating greater volatility. 

\setlength{\tabcolsep}{5pt}
\begin{table}[H]
\caption{Descriptive Statistics} 
\label{tab:general-statistics}
{\footnotesize
\begin{tabularx}{\textwidth}{@{}Xccccccc@{}}
\toprule[1.1pt]
{Data Set} & {\# Monthly} & {Mean} & {STD} & {Min} & {Max} & {\# of} & {Avg. Measurements} \\ 
 & Measurements & & & & & Indexes & per Index \\
 \cmidrule{2-8}
US - Headline Only     & 476 & 0.23 & 0.32 & -1.93 & 1.31 & 1 & 476 \\ 
US - Level 1           & 2064 & 0.19 & 0.85 & -18.61 & 11.32 & 26 & 79.38 \\ 
US - Level 2           & 7128 & 0.19 & 1.64 & -32.92 & 16.67 & 24 & 297 \\ 
US - Level 3           & 6476 & 0.17 & 1.75 & -34.24 & 24.81 & 25 & 259.04 \\ 
US - Level 4           & 9767 & 0.13 & 1.67 & -35.00 & 28.17 & 48 & 203.48 \\ 
US - Level 5           & 17656 & 0.11 & 2.39 & -23.89 & 242.50 & 110 & 160.51 \\ 
US - Level 6           & 10604 & 0.16 & 1.54 & -16.49 & 17.06 & 73 & 145.26 \\ 
US - Level 7           & 4968 & 0.21 & 1.64 & -11.89 & 17.84 & 36 & 138 \\ 
US - Level 8           & 980 & 0.21 & 1.70 & -8.65 & 7.80 & 7 & 140 \\ 
\bottomrule
\end{tabularx}\vspace{20pt}
\begin{tabularx}{\textwidth}{@{}Xccccccc@{}}
\toprule[1.1pt]
{Data Set} & {\# Monthly} & {Mean} & {STD} & {Min} & {Max} & {\# of} & {Avg. Measurements} \\ 
 & Measurements & & & & & Indexes & per Index \\
 \cmidrule{2-8}
Canada - Headline Only & 121 & 0.20 & 0.40 & -0.72 & 1.42 & 1 & 121 \\ 
Canada - Level 1       & 2057 & 0.19 & 1.11 & -10.78 & 8.17 & 17 & 121 \\ 
Canada - Level 2       & 2541 & 0.18 & 1.56 & -14.36 & 16.20 & 21 & 121 \\ 
Canada - Level 3       & 7381 & 0.20 & 1.92 & -20.69 & 27.92 & 61 & 121 \\ 
Canada - Level 4       & 10648 & 0.17 & 2.29 & -27.09 & 31.86 & 88 & 121 \\ 
Canada - Level 5       & 9680 & 0.19 & 2.93 & -34.03 & 35.43 & 80 & 121 \\ 
Canada - Level 6       & 3025 & 0.20 & 2.48 & -19.12 & 20.03 & 25 & 121 \\ 
\bottomrule
\end{tabularx}\vspace{20pt}
\begin{tabularx}{\textwidth}{@{}Xccccccc@{}}
\toprule[1.1pt]
{Data Set} & {\# Monthly} & {Mean} & {STD} & {Min} & {Max} & {\# of} & {Avg. Measurements} \\ 
 & Measurements & & & & & Indexes & per Index \\
 \cmidrule{2-8}
Norway - Headline Only & 172 & 0.22 & 0.45 & -0.93 & 1.36 & 1 & 172 \\ 
Norway - Level 1       & 2064 & 0.21 & 1.49 & -12.82 & 10.88 & 12 & 172 \\ 
Norway - Level 2       & 6708 & 0.21 & 2.15 & -20.79 & 27.31 & 39 & 172 \\ 
\bottomrule
\end{tabularx}
\begin{tablenotes}
\item {\footnotesize \textit{Notes:} General statistics of the headline CPI and CPI per each level in the hierarchy across Canada, Norway, and the US.}
\end{tablenotes}}
\end{table}

\begin{figure}[H]
    \centering
    
    \subfloat[Canada CPI by Hierarchy Level]{
        \includegraphics[width=0.6\textwidth]{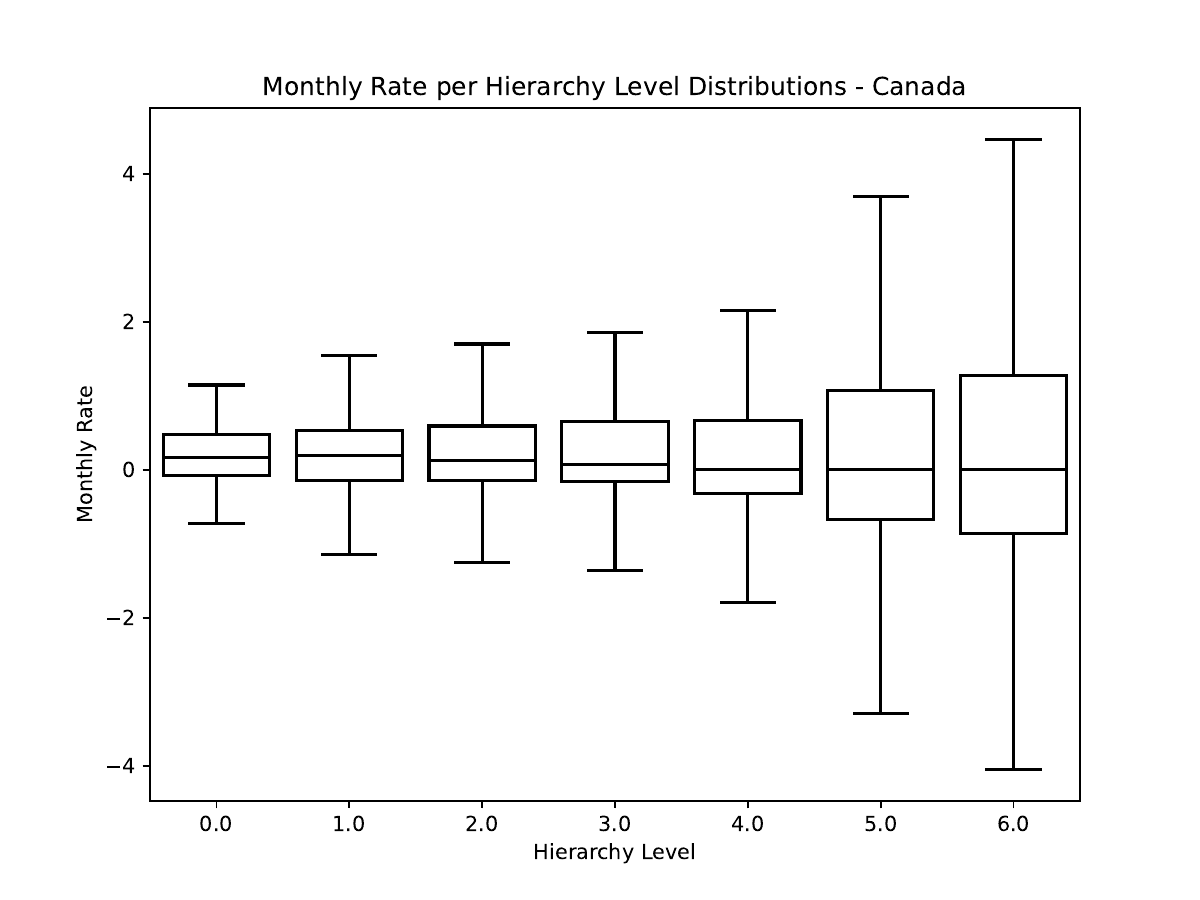}
        \label{fig:boxplots_hierarchy_canada}
    }
    \hfill
    \subfloat[Norway CPI by Hierarchy Level]{
        \includegraphics[width=0.6\textwidth]{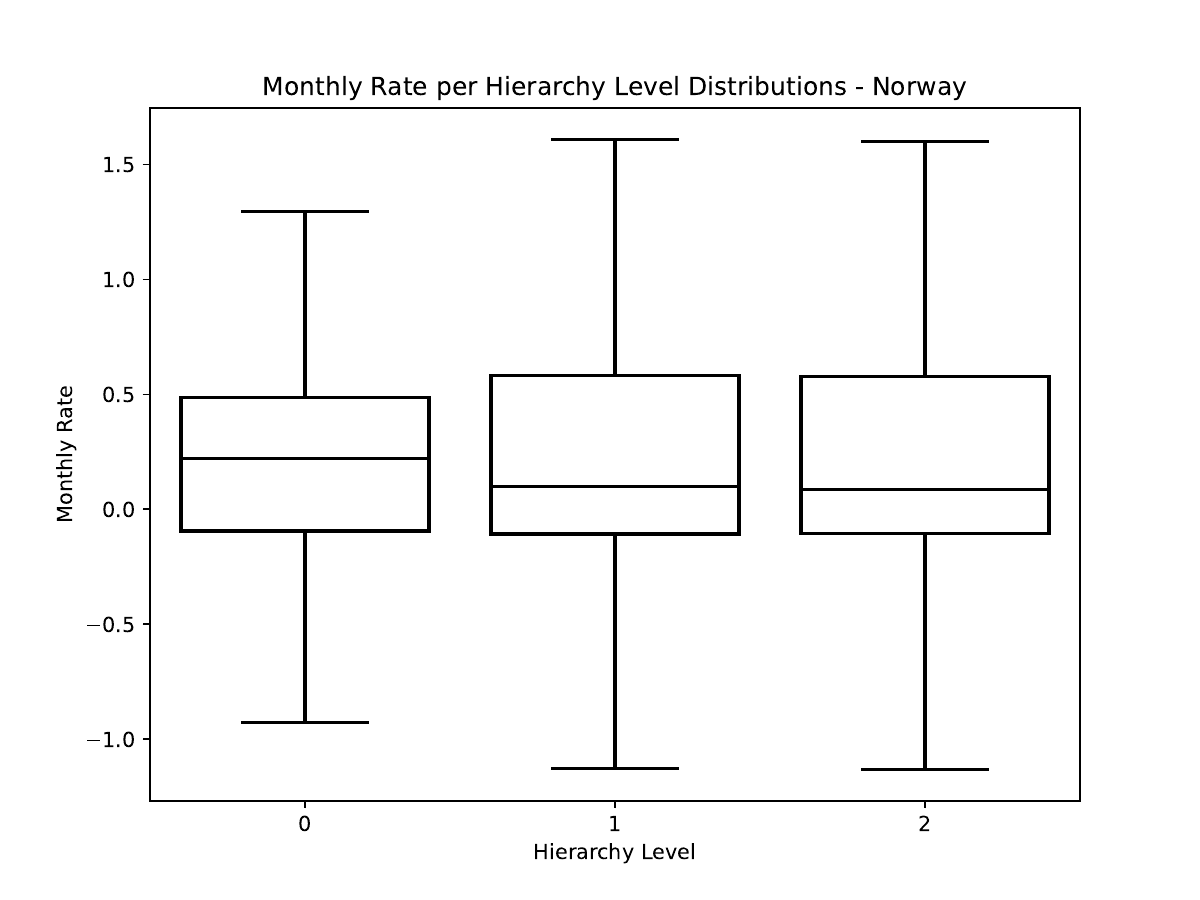}
        \label{fig:boxplots_hierarchy_norway}
    }
    \hfill
    \subfloat[US CPI by Hierarchy Level]{
        \includegraphics[width=0.6\textwidth]{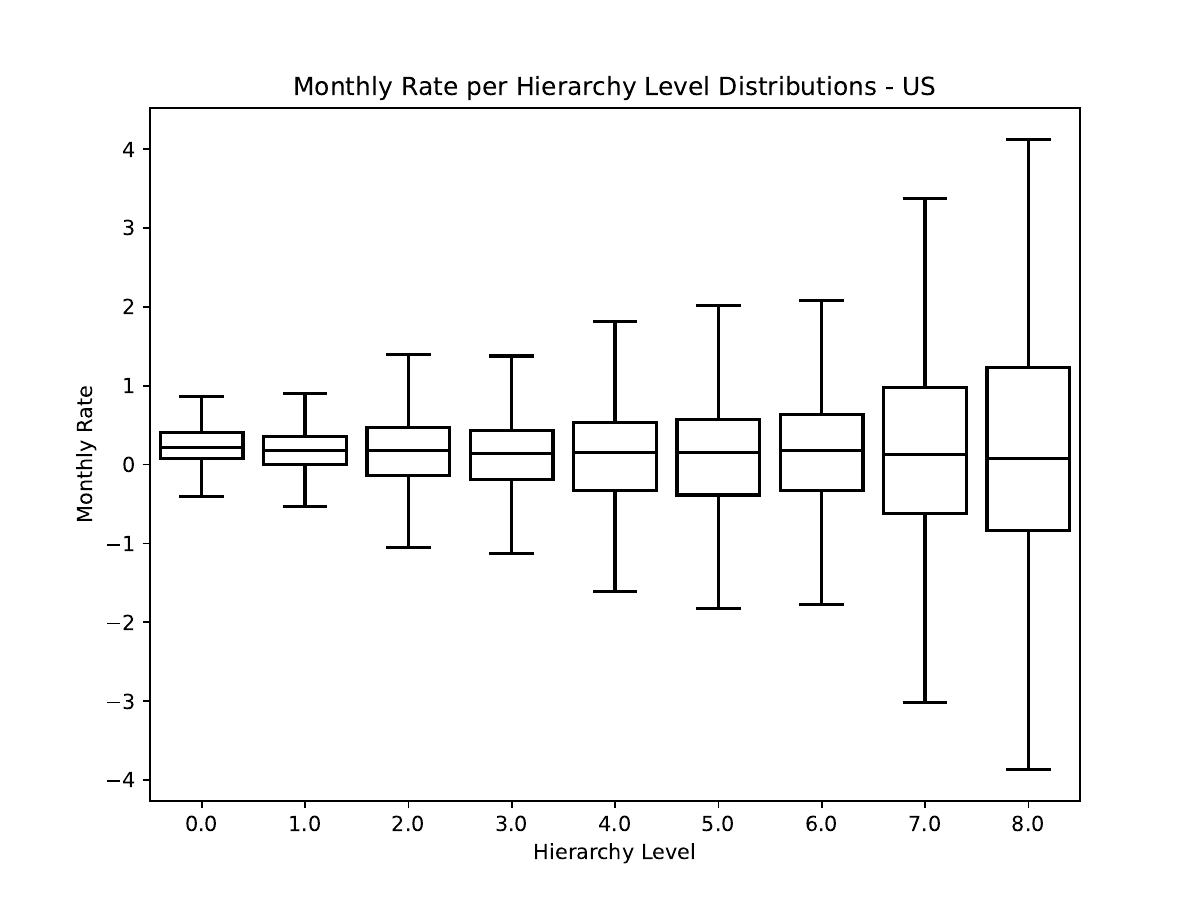}
        \label{fig:boxplots_hierarchy_us}
    }
    
    \caption{CPI Distributions by Hierarchy Level for Canada, Norway, and the US}
    \label{fig:boxplots_hierarchy_all}
\end{figure}

Figure~\ref{fig:boxplots_hierarchy_all} depicts box plots of the CPI change rate distributions at different levels. The boxes depict the median value and the upper 75'th and lower 25'th percentiles. The figure further emphasize that the change rates are more volatile as we go down the CPI hierarchy.
A high dynamic range, high standard deviation, and limited training data all signal increased difficulty in making predictions within the hierarchy. Given this, we can anticipate that predictions for disaggregated components within the hierarchy will be more challenging than those for the headline.

\begin{figure}[H]
    \centering
    \includegraphics[width=1.1\textwidth]{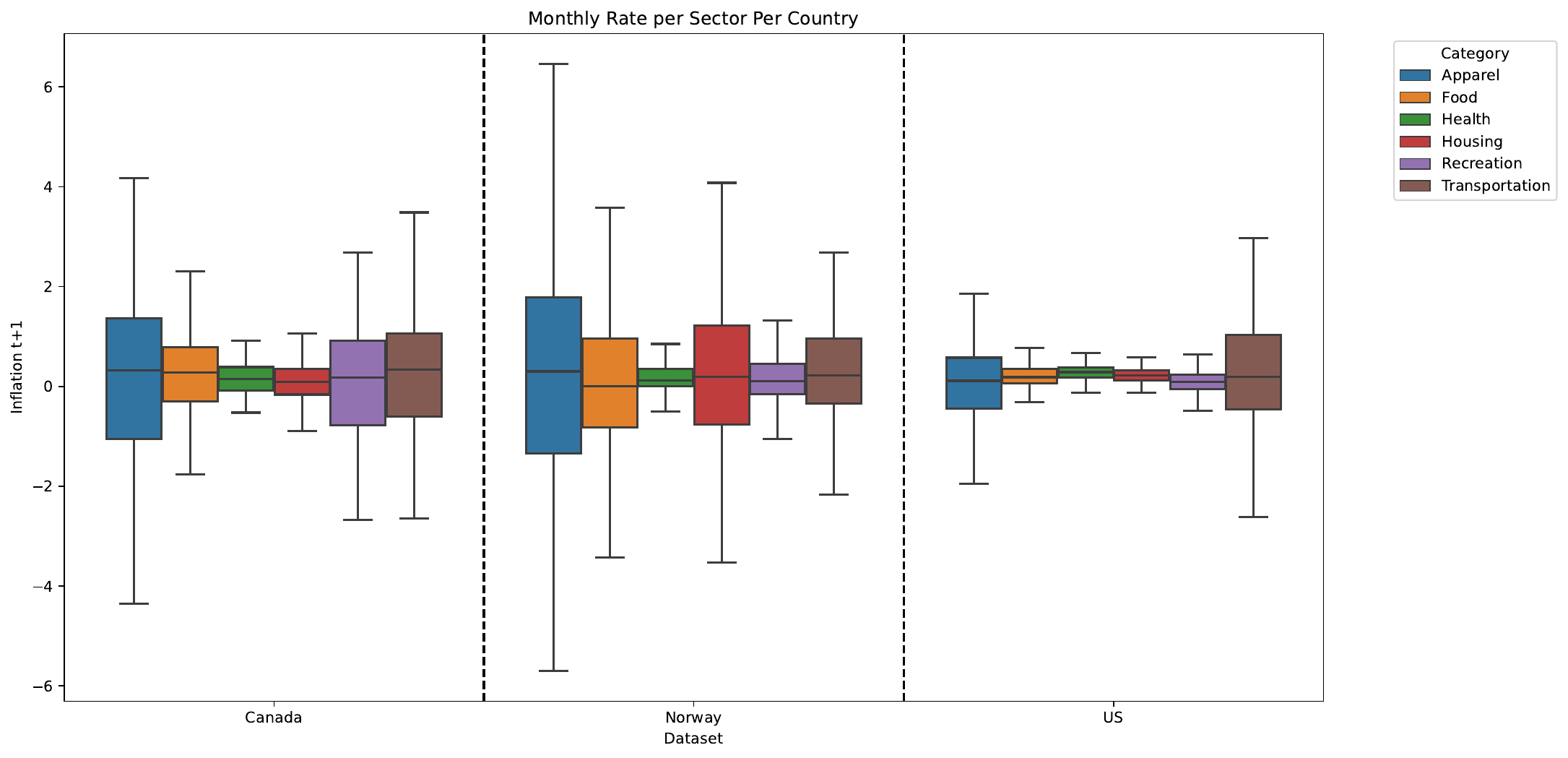} 
    \caption{Monthly Rate per Sector Per Country}
    \label{fig:boxplot_per_sector}
\end{figure}

Figure~\ref{fig:boxplot_per_sector} presents a box plot showing the distribution of CPI change rates across various sectors. It is evident that certain sectors, like Apparel and Transportation, exhibit greater volatility than others across all markets. This higher volatility is likely to make predictions for these sectors more challenging, as anticipated.

\subsubsection{Data Augmentation}\label{subsec:data_aug}
As part of the pre-processing process we addressed the issue of missing weights in the Canadian dataset by employing a linear regression-based imputation method. In this approach, the prices of child categories were used as the independent variables (predictors), while the price of the parent category was treated as the dependent variable (target). This regression model allowed us to estimate the missing weights by leveraging the relationship between the prices of child and parent categories, thereby ensuring a consistent and reliable imputation strategy. Since the model's loss function incorporates the weights of each child category, this imputation step was a critical part of the preprocessing pipeline.

\chapter{Evaluation and Results} 
We evaluate  the performance of the BiHRNN and compare it to well-established baselines for inflation prediction, as well as some additional machine learning approaches.
We use the following notation: Let $x_t$ be the CPI log-change rate at month $t$ .
Models for $\hat{x}_{t}$ are considered as an estimate for $x_t$ based on historical values.
Furthermore, we denote the estimation error at time $t$ by $\varepsilon_{t}$ .
Hyper-parameters were set using Bayesian-inspired optimization procedures.

\section{Baseline Models} \label{baslines}
We compare Bidirectional HRNN with the following CPI prediction baselines:

\begin{enumerate}

\item{\bf Autoregression (AR) -} The AR($\rho$) model estimates $\hat{x}_{t}$ based on the previous $\rho$ timestamps using the following equation: $\hat{x}_{t}= \alpha_0 + \left(\sum_{i=1}^{\rho} \alpha_{i} x_{t-i} \right) + \varepsilon_{t}$, where $\{ \alpha_i \}_{i=0}^{\rho}$ represent the model's parameters.

\item{\bf Random Walk (RW) -} We consider the RW($\rho$) model from \cite{atkeson2002}. RW($\rho$) is a straightforward but powerful model that predicts the next timestamps by taking the average of the last $\rho$ timestamps, using the formula:  $\hat{y}_{t}=\frac{1}{\rho} \sum_{i=1}^{\rho} x_{t-i} + \varepsilon_{t}$.

\item{\bf Random Forests (RF) - } The RF($\rho$) model is an ensemble learning approach that constructs multiple decision trees~\citep{song2015decision} to reduce overfitting and enhance generalization~\citep{breiman2001random}. During prediction, the model returns the average of the predictions made by each individual tree. The inputs to the RF($\rho$) model are the last $\rho$ samples, and the output is the predicted value for the next timestamp.

\item{\bf Extreme Gradient Boost (XGBoost) - } The XGBoost($\rho$) model \citep{Chen_2016} is based on an ensemble of decision trees which are trained in a stage-wise fashion similar to other boosting models \citep{schapire1999brief}. Unlike RF($\rho$) which averages the prediction of multiple decision trees, the XGBoost($\rho$) trains each tree to minimize the remaining residual error of all previous trees. At prediction time, the sum of predictions of all the trees is returned.  The inputs to the XGBoost($\rho$) model are the last $\rho$ samples and the output is the predicted value for the next timestamps.

\item{\bf Fully Connected Neural Network (FC) -} The FC($\rho$) model is a fully connected neural network with one hidden layer of size 100 and a ReLU activation function~\citep{ActivationFunctions}. The output layer does not use any activation function to frame the task as a regression problem, optimized using a squared loss function. The inputs to the FC($\rho$) model consist of the last $\rho$ samples, and the output is the predicted value for the next timestamp.

\item{\bf Support Vector Regression (SVR) - } 
SVR($\rho$)  is a machine learning model based on Support Vector Machines (SVM), used for regression tasks ~\cite{NIPS1996_d3890178}.  SVR($\rho$) attempts to find a function that fits the data within a certain margin of tolerance, while minimizing the prediction error outside this margin. It is particularly effective for capturing complex relationships in data and is robust to outliers due to its focus on maximizing the margin around the prediction.  The kernel used for the prediction is "rbf" and the degree of the polynomial kernel function is three.
The inputs to the SVR($\rho$) model are the last $\rho$ samples and the output is the predicted value for the next timestamps.
\end{enumerate}

\section{Ablation Models}
To highlight the impact of the hierarchical component in the Bidirectional HRNN model, we performed an ablation study by comparing it to "simpler" alternatives, specifically GRU-based models that exclude the hierarchical component: 

\begin{enumerate}

\item{\bf Single GRU (S-GRU) -} The S-GRU($\rho$) is a single GRU unit that receives the last $\rho$ values as inputs in order to predict the next value. In GRU($\rho$), a single GRU is used for all the time series that comprise the CPI hierarchy. This baseline utilizes all the benefits of a GRU but assumes that the different components of the CPI behave similarly and a single unit is sufficient to model all the nodes.   

\item {\bf Independent GRUs (I-GRUs) -}
In I-GRUs($\rho$), we trained a different GRU($\rho$) unit for each CPI node. 
The S-GRU and I-GRU approaches represent two extremes: The first attempts to model all the CPI nodes with a single model, while the second treats each node separately. 

To emphasize the effect of bidirectionality, we also compared the Bidirectional HRNN to its predecessor, the Hierarchical Recurrent Neural Network (HRNN).

\item {\bf Hierarchical Recurrent Neural Network (HRNN) -}
In HRNN($\rho$), we trained a separate GRU($\rho$) unit for each CPI node, while incorporating the model weights of its parent category. This approach allows information to flow from parent to child categories, effectively leveraging the hierarchical structure of the data and enhancing prediction accuracy.
\end{enumerate}

\section{Evaluation Metrics}
Following \cite{faust2013forecasting} and \cite{AparicioBertolotto2020a}, we report our results using three evaluation metrics: 
\begin{enumerate}
    \item{\bf Root Mean Squared Error (RMSE) -} 
    The RMSE is calculated as: \begin{equation}
        RMSE=\sqrt{\frac{1}{T}\sum_{t=1}^T \left(x_t- \hat{x}_t \right)^2},
    \end{equation}
     where $x_t$ represents the actual monthly change rate for month $t$, and $\hat{x}_t$ denotes the corresponding predicted value.
    
    \item{\bf Pearson Correlation Coefficient -} The Pearson correlation coefficient $\phi$ is defined as:
    \begin{equation}
        \phi = \frac{COV(X_T,\hat{X}_T)}{\sigma_{X} \times \sigma_{\hat{X}}},    
    \end{equation}
        where $COV(X_T,\hat{X}_T)$ is the covariance between the actual values and predictions, and $\sigma_{X_T}$ and $\sigma_{\hat{X}_T}$ are the standard deviations of the actual values and the predictions, respectively.
    
\item{\bf Distance Correlation Coefficient -} 
Unlike the Pearson correlation, which only measures linear relationships, the distance correlation coefficient can detect both linear and nonlinear associations ~\citep{SzekelyRizzoBakirov2007a,distanceCorrelation}. 
The distance correlation coefficient $r_d$ is given by:
    \begin{equation}
    \label{eq:distance_corr}
        r_d= \frac{\operatorname{dCov}(X_T, \hat{X}_T)}{\sqrt{ \operatorname{dVar}(X_T) \times \operatorname{dVar}(\hat{X}_T)}}
    \end{equation}
where $\operatorname{dCov}(X_T, \hat{X}_T)$ is the distance covariance between the actual values and the predictions, and $\operatorname{dVar}(X_T)$ and $\operatorname{dVar}(\hat{X}_T)$ are the distance variances of the actual values and the predictions, respectively.








\end{enumerate}

\section{Results} 
\label{sec:Results}
The BiHRNN model stands out for its ability to leverage information flow both from higher levels to lower levels and from lower levels to higher levels within the hierarchy. The model leverages the inherent hierarchy of the CPI, enhancing predictions at both granular and broader, more significant levels, such as the CPI Headline.
Therefore, we will provide the headline results separately, along with the aggregated results across all categories.

The results are relative to the $AR(1)$ model and normalized according to: $\frac{RMSE_{Model}}{RMSE_{AR\left( 1\right) }}$.

\subsection{US CPI Results}
\setlength{\tabcolsep}{3pt}
\begin{table}[H]
\begin{threeparttable} 
\caption{Average Results on Disaggregated CPI Components - US} 
\label{tab:allCPIResults - US}
{\scriptsize  
\begin{tabularx}{\textwidth}{l>{\centering\arraybackslash}X>{\centering\arraybackslash}X>{\centering\arraybackslash}X>{\centering\arraybackslash}X>{\centering\arraybackslash}X>{\centering\arraybackslash}X>{\centering\arraybackslash}X>{\centering\arraybackslash}X}
\toprule[1.1pt]
\textbf{Model} & \textbf{\parbox[c]{1cm}{\centering Avg. \\ RMSE}}  & \textbf{\parbox[c]{1.2cm}{\centering Pearson \\ Corr.}} & \textbf{\parbox[c]{1.2cm}{\centering Dist. \\ Corr.}} & \textbf{\parbox[c]{1.2cm}{\centering Headline \\ RMSE}} & \textbf{\parbox[c]{1.4cm}{\centering Headline Pearson \\ Corr.}} & \textbf{\parbox[c]{1.4cm}{\centering Headline Dist. \\ Corr.}} \\ 
\midrule
I-GRU & 1.215 & 0.138 & 0.338 & 1.015 & 0.347 & 0.350 \\
AR\_1 & 1.000 & 0.176 & 0.513 & 1.000 & 0.327 & 0.459 \\
AR\_2 & 1.267 & 0.105 & 0.467 & 1.312 & 0.327 & 0.565 \\
AR\_3 & 1.487 & 0.082 & 0.437 & 1.560 & 0.349 & 0.510 \\
AR\_4 & 1.902 & 0.052 & 0.411 & 1.749 & 0.308 & 0.427 \\
FC\_p\_12 & 1.229 & -0.014 & 0.355 & 1.592 & 0.027 & 0.251 \\
RF\_p\_12 & 1.143 & 0.112 & 0.377 & 1.210 & 0.368 & 0.377 \\
RW\_p\_4 & 1.189 & -0.013 & 0.353 & 1.420 & -0.050 & 0.310 \\
SVR\_p\_12 & 1.115 & 0.067 & 0.363 & 1.280 & 0.473 & 0.529 \\
XGB\_p\_12 & 1.228 & 0.087 & 0.369 & 1.312 & 0.392 & 0.423 \\
HRNN & 1.028 & 0.158 & 0.346 & 1.015 & 0.347 & 0.350 \\
BiHRNN & 0.966 & 0.230 & 0.378 & 1.052 & 0.225 & 0.290 \\
\bottomrule[1.1pt]
\end{tabularx}
}
\end{threeparttable}
\end{table}

\subsection{Canada CPI Results}
\setlength{\tabcolsep}{3pt}
\begin{table}[H]
\begin{threeparttable} 
\caption{Average Results on Disaggregated CPI Components - Canada} 
\label{tab:allCPIResults - Canada}
{\scriptsize  
\begin{tabularx}{\textwidth}{l>{\centering\arraybackslash}X>{\centering\arraybackslash}X>{\centering\arraybackslash}X>{\centering\arraybackslash}X>{\centering\arraybackslash}X>{\centering\arraybackslash}X>{\centering\arraybackslash}X>{\centering\arraybackslash}X}
\toprule[1.1pt]
\textbf{Model} & \textbf{\parbox[c]{1cm}{\centering Avg. \\ RMSE}} & \textbf{\parbox[c]{1.2cm}{\centering Pearson \\ Corr.}} & \textbf{\parbox[c]{1.2cm}{\centering Dist. \\ Corr.}} & \textbf{\parbox[c]{1.2cm}{\centering Headline \\ RMSE}} & \textbf{\parbox[c]{1.4cm}{\centering Headline Pearson \\ Corr.}} & \textbf{\parbox[c]{1.4cm}{\centering Headline Dist. \\ Corr.}} \\ 
\midrule
I-GRU & 0.892 & 0.351 & 0.476 & 1.261 & 0.329 & 0.516 \\
AR\_1 & 1.000 & 0.128 & 0.633 & 1.000 & 0.628 & 0.693 \\
AR\_2 & 1.188 & -0.008 & 0.592 & 1.155 & 0.241 & 0.409 \\
AR\_3 & 1.305 & -0.004 & 0.561 & 0.970 & 0.646 & 0.671 \\
AR\_4 & 1.684 & -0.004 & 0.522 & 1.424 & 0.356 & 0.416 \\
FC\_p\_12 & 0.907 & 0.237 & 0.465 & 2.051 & 0.152 & 0.352 \\
RF\_p\_12 & 0.861 & 0.318 & 0.509 & 1.635 & 0.542 & 0.540 \\
RW\_p\_4 & 0.901 & 0.138 & 0.414 & 1.261 & 0.495 & 0.658 \\
SVR\_p\_12 & 0.852 & 0.308 & 0.517 & 1.721 & 0.449 & 0.580 \\
XGB\_p\_12 & 0.936 & 0.271 & 0.502 & 1.635 & 0.411 & 0.527 \\
HRNN & 0.824 & 0.358 & 0.490 & 1.261 & 0.329 & 0.516 \\
BiHRNN & 0.795 & 0.386 & 0.511 & 1.170 & 0.321 & 0.497 \\
\bottomrule[1.1pt]
\end{tabularx}
}
\end{threeparttable}
\end{table}

\subsection{Norway CPI Results}
\setlength{\tabcolsep}{3pt}
\begin{table}[H]
\begin{threeparttable} 
\caption{Average Results on Disaggregated CPI Components - Norway} 
\label{tab:allCPIResults - Norway}
{\scriptsize  
\begin{tabularx}{\textwidth}{l>{\centering\arraybackslash}X>{\centering\arraybackslash}X>{\centering\arraybackslash}X>{\centering\arraybackslash}X>{\centering\arraybackslash}X>{\centering\arraybackslash}X>{\centering\arraybackslash}X>{\centering\arraybackslash}X}
\toprule[1.1pt]
\textbf{Model} & \textbf{\parbox[c]{1cm}{\centering Avg. \\ RMSE}} & \textbf{\parbox[c]{1.2cm}{\centering Pearson \\ Corr.}} & \textbf{\parbox[c]{1.2cm}{\centering Dist. \\ Corr.}} & \textbf{\parbox[c]{1.2cm}{\centering Headline \\ RMSE}} & \textbf{\parbox[c]{1.4cm}{\centering Headline Pearson \\ Corr.}} & \textbf{\parbox[c]{1.4cm}{\centering Headline Dist. \\ Corr.}} \\ 
\midrule
I-GRU & 0.866 & 0.355 & 0.5063 & 0.724 & 0.208 & 0.413 \\
AR\_1 & 1.000 & 0.053 & 0.653 & 1.000 & -0.583 & 0.546 \\
AR\_2 & 1.251 & -0.003 & 0.617 & 1.132 & -0.743 & 0.695 \\
AR\_3 & 1.378 & 0.009 & 0.568 & 1.189 & -0.797 & 0.777 \\
AR\_4 & 1.727 &  0.011 & 0.535 & 1.296 & -0.536 & 0.574 \\
FC\_p\_12 & 0.974 & 0.226 & 0.454 & 0.924 & 0.242 & 0.422 \\
RF\_p\_12 & 0.849 & 0.349 & 0.539 & 0.672 & 0.613 & 0.721 \\
RW\_p\_4 & 0.973 & 0.187 & 0.405 & 0.788 & 0.165 & 0.337 \\
SVR\_p\_12 & 0.851 & 0.376 & 0.550 & 0.848 & 0.251 & 0.362 \\
XGB\_p\_12 & 0.904 & 0.303 & 0.530 & 0.669 & 0.644 & 0.725 \\
HRNN & 0.832 & 0.3677 & 0.5365 & 0.724 & 0.208 & 0.413 \\
BiHRNN & 0.767 & 0.478 & 0.567 & 0.655 & 0.390 & 0.477 \\
\bottomrule[1.1pt]
\end{tabularx}
}
\end{threeparttable}
\end{table}

The results in Tables ~\ref{tab:allCPIResults - Canada}, ~\ref{tab:allCPIResults - Norway}, and ~\ref{tab:allCPIResults - US} show that the BiHRNN consistently outperforms other models in terms of predictive accuracy and stability. With some of the lowest RMSE values across datasets, this model demonstrates its ability to reliably minimize error between various components of the CPI. In comparison, simpler models like AR(1) and AR(4) often exhibit higher RMSE and greater variability, indicating that the BiHRNN model offers a stronger, more stable fit for this complex data.


Correlation metrics reinforce this model’s capacity to understand the underlying relationships in the data. The BiHRNN achieves high Pearson and Distance correlations indicating a strong alignment between model predictions and actual outcomes. Although a few models, like RF(12), show competitive correlations, their higher RMSE values demonstrate an inability to consistently maintain accuracy across metrics.

The BiHRNN model demonstrates top-tier performance in headline predictions, excelling in both RMSE and correlation metrics. However, our findings indicate that the Headline data alone is sufficient for accurate headline predictions and yields the best results. Attempts to incorporate additional regularization terms do not enhance prediction performance. Consequently, we recommend that future work on headline predictions focus exclusively on using the Headline data.

This balance across both disaggregated components and headline metrics highlights the model's robustness and adaptability, making it a preferable choice for forecasting CPI trends. Overall, the BiHRNN stands out as the most effective model, combining low error rates, strong fit, and high correlation, all of which contribute to a more accurate and reliable CPI prediction framework across categories.

Figure ~\ref{fig:disaggregated_index_predictions} below showcases examples of several disaggregated indexes from different hierarchy levels and sectors. The solid black line shows the actual CPI values, while the dashed lines depict predictions from the top-performing models—all variations of RNN models: BiHRNN, HRNN, and I-GRU in blue, green, and red, respectively. As shown in the graphs, the BiHRNN model demonstrates superior predictive accuracy, achieving lower RMSEs and more effectively capturing shifts in trends compared to its counterparts.

\begin{figure}[H]
    \centering
    
    \subfloat[Food]{
        \includegraphics[width=0.6\textwidth]{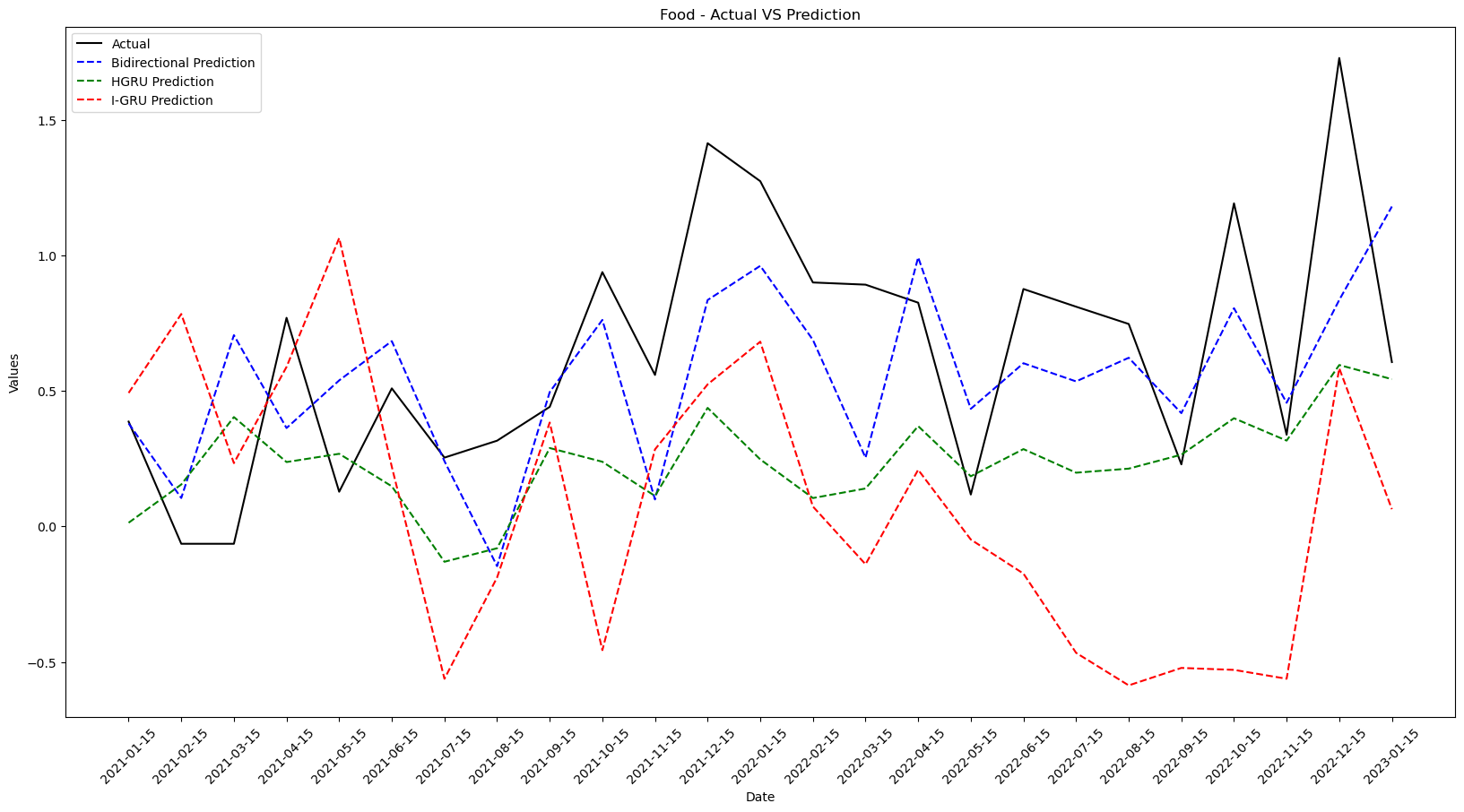}
        \label{fig:Food - Canada}
    }
    \hfill
    \subfloat[Housekeeping]{
        \includegraphics[width=0.6\textwidth]{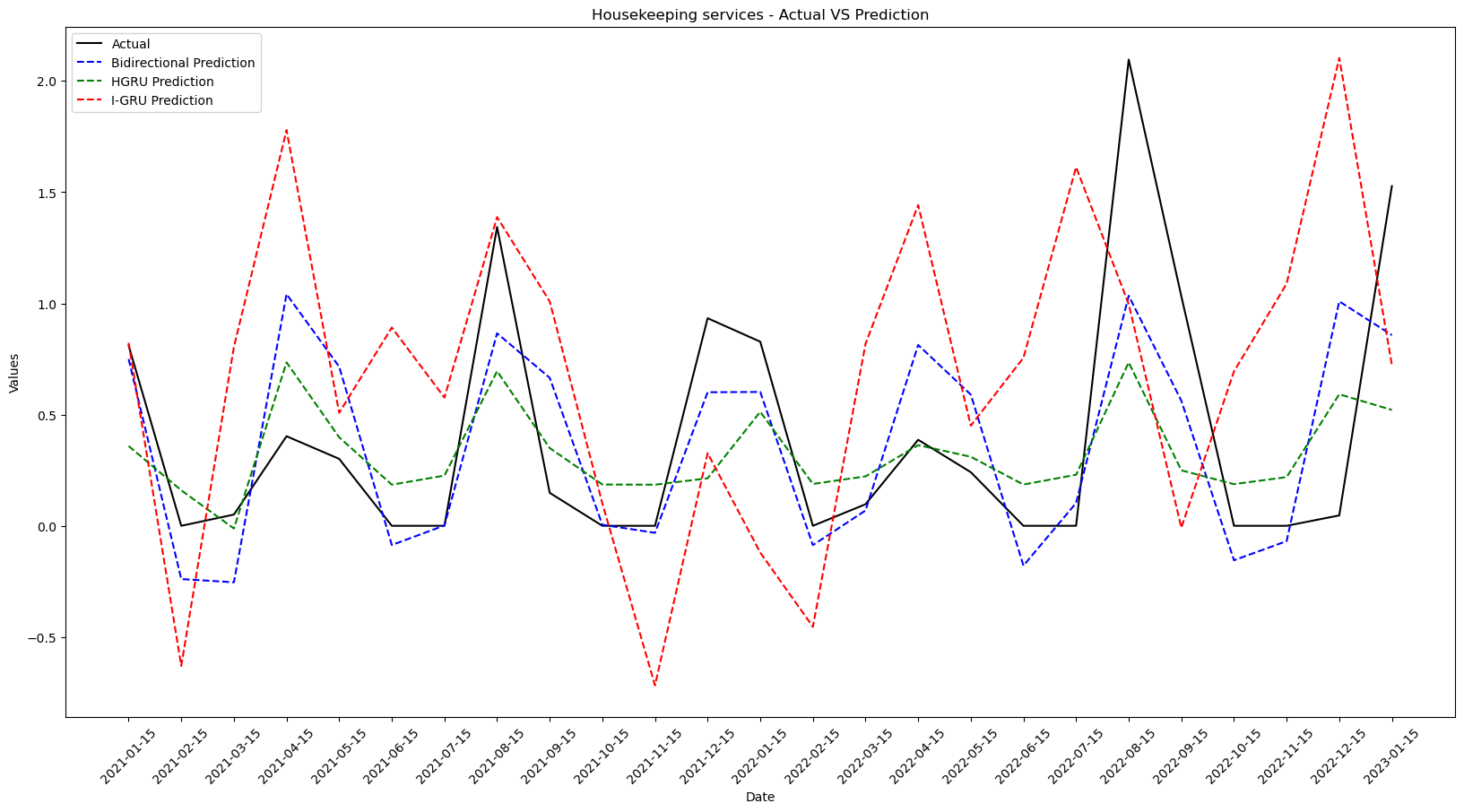}
        \label{fig:Housekeeping - Canada}
    }
    
    
    \subfloat[Footwear]{
        \includegraphics[width=0.6\textwidth]{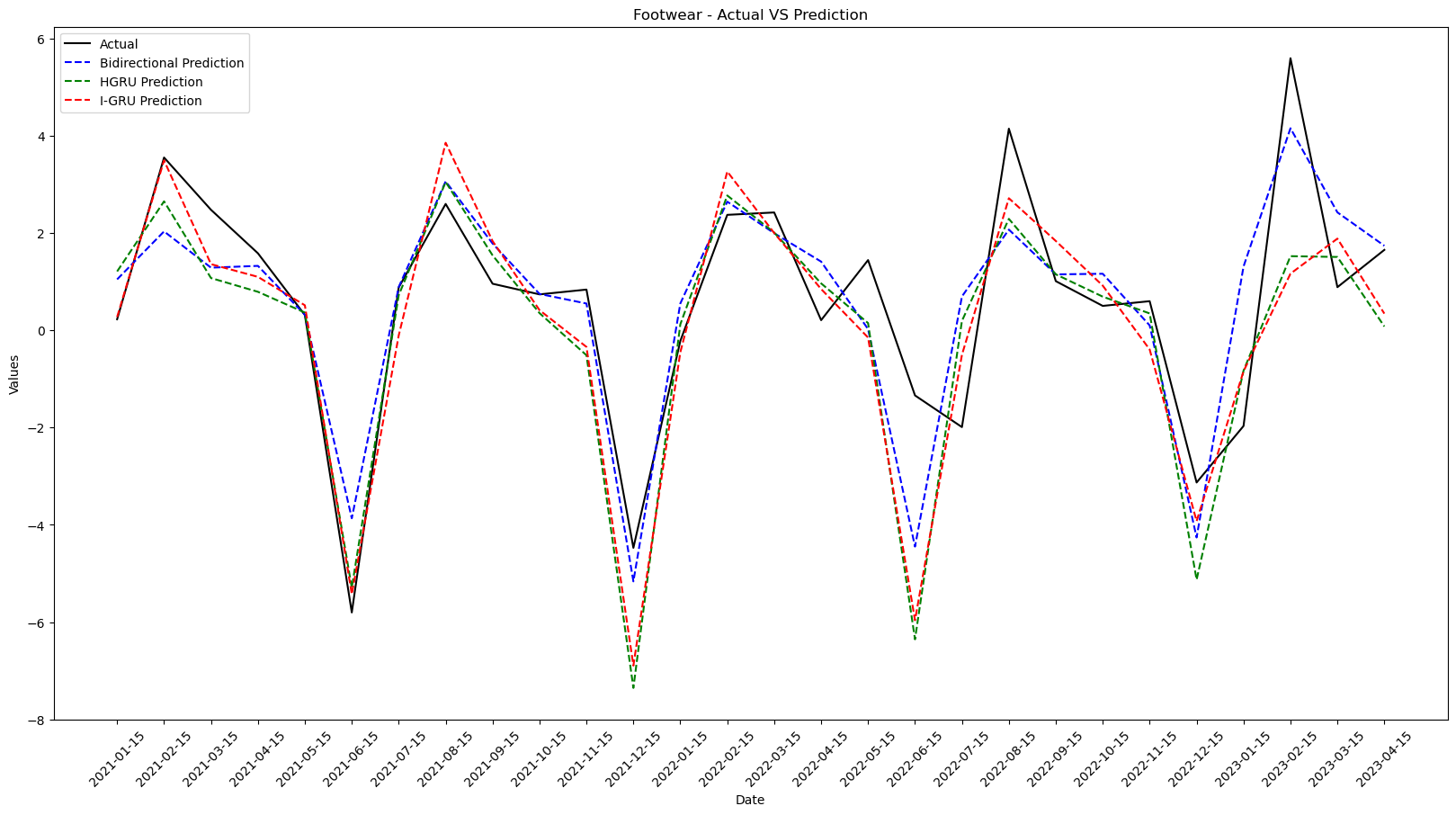}
        \label{fig:Footwear Norway}
    }
    \hfill
    \subfloat[Out-patient Services]{
        \includegraphics[width=0.6\textwidth]{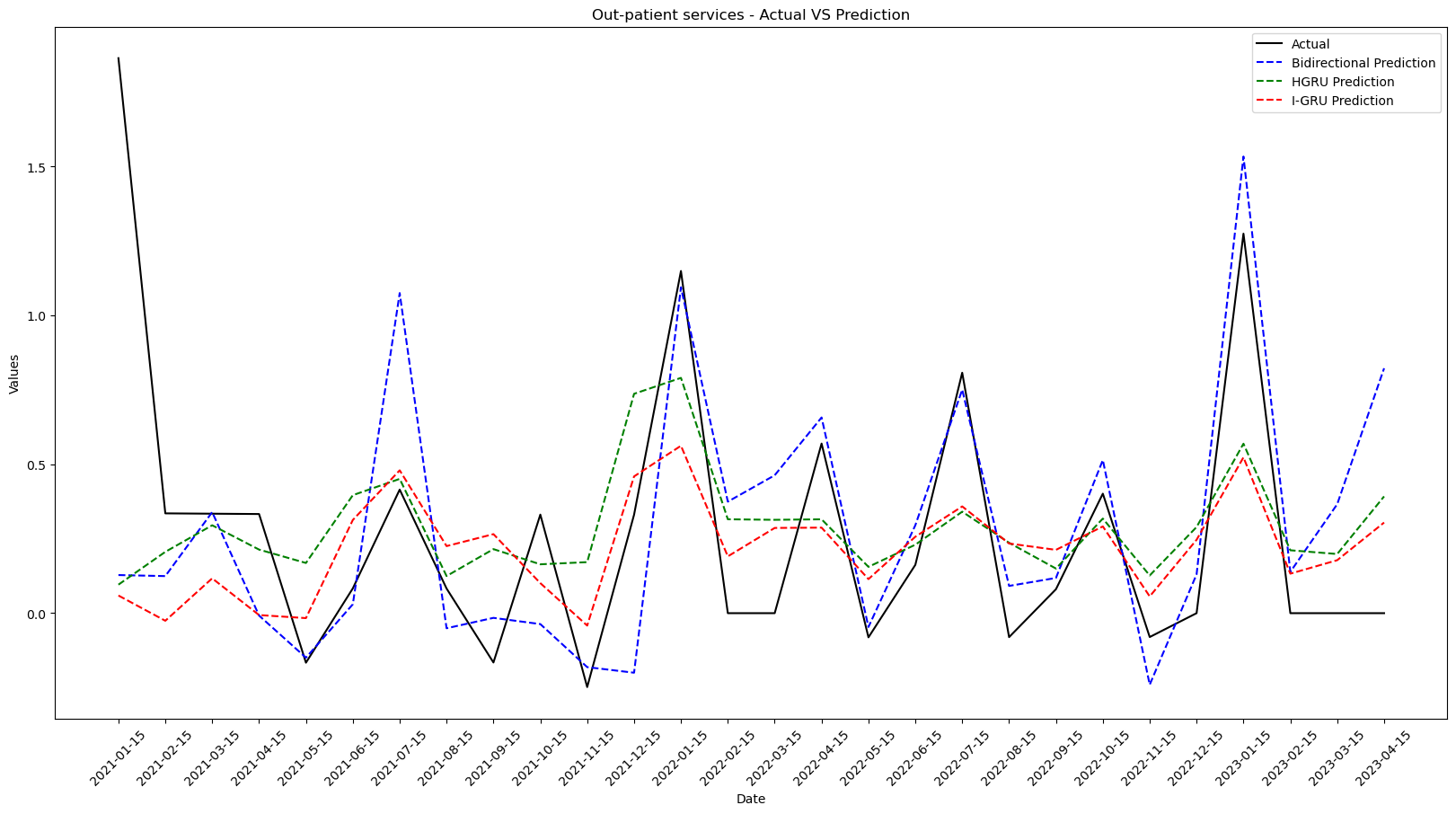}
        \label{fig:Out-patient services - Norway}
    }
    
    \caption{Disaggregated Index CPI Predictions}
    \label{fig:disaggregated_index_predictions}
\end{figure}

\chapter{Conclusion}

This thesis presented a novel approach to inflation time series forecasting using Bidirectional Hierarchical Recurrent Neural Networks (BiHRNNs). The primary objective was to leverage hierarchical data structures to enhance prediction accuracy at both granular and aggregate levels of Consumer Price Index (CPI) data. The proposed BiHRNN model was applied to CPI data from three established markets: Canada, Norway, and the United States.

The BiHRNN model was trained and evaluated on these datasets, with hyperparameters carefully optimized for each market. Additionally, we experimented with loss functions designed to propagate information between hierarchical levels, which proved crucial for achieving superior forecasting accuracy. Compared to traditional RNNs and machine learning models, the BiHRNN demonstrated significantly improved performance across all hierarchical levels.

Future work will extend the application of BiHRNNs to CPI datasets from other countries and sectors, addressing the unique challenges they present. Further enhancements will focus on incorporating additional features and exploring advanced methods for parameter optimization and loss function design.

In summary, this thesis established BiHRNNs as a robust framework for inflation time series forecasting, highlighting their ability to surpass traditional methods and laying the groundwork for further research and innovation in this area.

\bibliographystyle{plainnat}
\bibliography{references}


\newpage{}

\includepdf[pages=-]{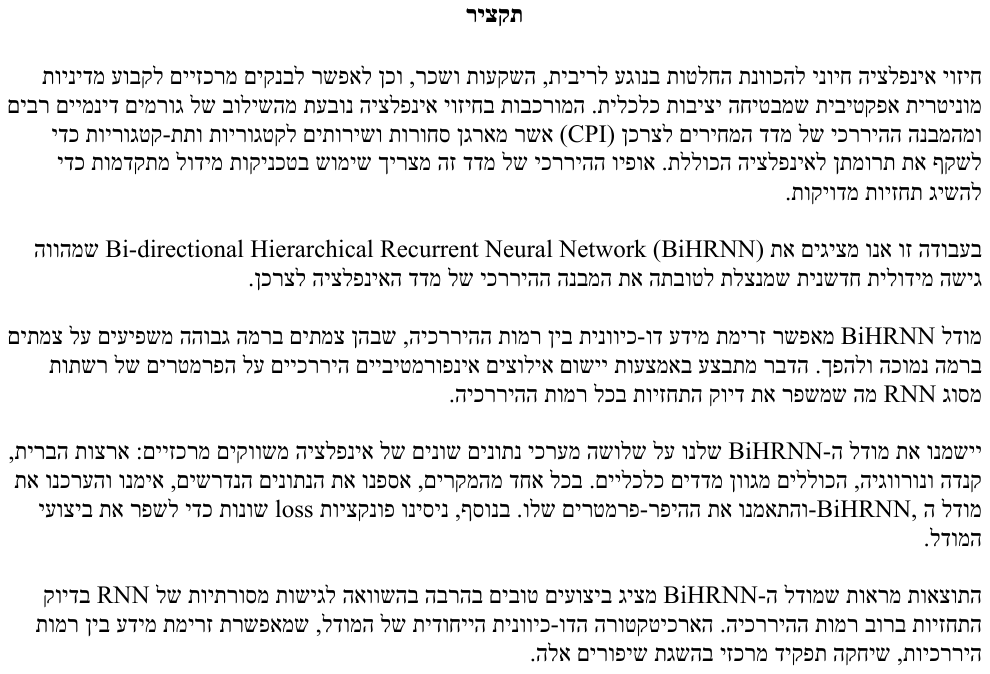}
\end{document}